\documentclass{article}
\usepackage[arxiv]{colm2026_conference}

\usepackage{microtype}
\usepackage{amsmath}
\usepackage{hyperref}
\usepackage{cleveref}
\usepackage{url}
\usepackage{booktabs}
\usepackage{graphicx}
\usepackage{wrapfig}
\usepackage{subcaption}
\usepackage{subcaption}

\usepackage{xcolor}
\usepackage[most,skins,theorems]{tcolorbox}
\usepackage{tikz}
\usepackage{lineno}
\usepackage[pdftex,dvipsnames]{xcolor}  
\usepackage[colorinlistoftodos,prependcaption,textsize=tiny]{todonotes}
\usepackage[colorinlistoftodos]{todonotes}
\usepackage{enumitem}

\crefname{section}{Section}{Sections}
\crefname{table}{Table}{Tables}
\crefname{figure}{Figure}{Figures}

\crefrangeformat{section}{\S#3#1#4\crefrangeconjunction\S#5#2#6}
\crefmultiformat{section}{\S#2#1#3}{\crefpairconjunction\S#2#1#3}{\crefmiddleconjunction\S#2#1#3}{\creflastconjunction\S#2#1#3}
\newcommand{\crefrangeconjunction}{--}
\crefname{listing}{Lst.}{listings}
\crefname{line}{Lin.}{Lin.}
\crefname{appendix}{App.}{App.}
\usetikzlibrary{arrows.meta,positioning,fit}

\definecolor{darkblue}{rgb}{0, 0, 0.5}
\definecolor{promptbg}{RGB}{232,242,255}
\definecolor{promptframe}{RGB}{92,140,214}
\definecolor{evalbg}{RGB}{241,252,241}
\definecolor{evalframe}{RGB}{77,162,94}
\definecolor{solutionbg}{RGB}{255,246,232}
\definecolor{solutionframe}{RGB}{214,139,42}
\hypersetup{colorlinks=true, citecolor=darkblue, linkcolor=darkblue, urlcolor=darkblue}

\newtcolorbox{promptbox}[1]{
  enhanced,
  breakable,
  arc=2.2mm,
  boxrule=0.8pt,
  left=1.3mm,right=1.3mm,top=1.0mm,bottom=1.1mm,
  colback=promptbg,
  colframe=promptframe,
  colbacktitle=promptframe!34,
  coltitle=black,
  fonttitle=\bfseries,
  title={#1}
}

\newtcolorbox{evalbox}[1]{
  enhanced,
  breakable,
  arc=2.2mm,
  boxrule=0.8pt,
  left=1.3mm,right=1.3mm,top=1.0mm,bottom=1.1mm,
  colback=evalbg,
  colframe=evalframe,
  colbacktitle=evalframe!30,
  coltitle=black,
  fonttitle=\bfseries,
  title={#1}
}

\newtcolorbox{smallevalbox}[1][]{
  enhanced,
  breakable,
  arc=2.2mm,
  boxrule=0.8pt,
  left=1.3mm,right=1.3mm,top=1.0mm,bottom=1.1mm,
  colback=evalbg,
  colframe=evalframe,
  colbacktitle=evalframe!30,
  coltitle=black,
  fonttitle=\bfseries,
  title={#1}
}

\newtcolorbox{solutionbox}[1]{
  enhanced,
  breakable,
  arc=2.2mm,
  boxrule=0.8pt,
  left=1.3mm,right=1.3mm,top=1.0mm,bottom=1.1mm,
  colback=solutionbg,
  colframe=solutionframe,
  colbacktitle=solutionframe!34,
  coltitle=black,
  fonttitle=\bfseries,
  title={#1}
}

\definecolor{rliableolive}{HTML}{BBCC33}
\tcbset{
  aibox/.style={
    width=\linewidth,
    top=7pt,
    bottom=2pt,
    colback=rliableolive!13!white,
    colframe=black,
    colbacktitle=black,
    enhanced,
    center,
    attach boxed title to top left={yshift=-0.1in,xshift=0.15in},
    boxed title style={boxrule=0pt,colframe=white,},
  }
}

\newtcolorbox{AIbox}[2][]{aibox,title=#2,#1}

\title{QED-Nano: Teaching a Tiny Model to Prove Hard Theorems}

\author{LM Provers Team$^{1,2,3,4}$\\
$^1$CMU: Yuxiao Qu, Amrith Setlur, Ian Wu, Aviral Kumar\\
$^2$Hugging Face: Edward Beeching, Lewis Tunstall\\
$^3$ETH Zurich: Jasper Dekoninck\\
$^4$Project Numina: Jia Li\\
\texttt{https://huggingface.co/lm-provers}}

\begin{document}

\ifcolmsubmission
\linenumbers
\fi

\maketitle

\begin{abstract}

Proprietary AI systems have recently demonstrated impressive capabilities on complex proof-based problems, with gold-level performance reported at the 2025 International Mathematical Olympiad (IMO). However, the training pipelines behind these systems remain largely undisclosed, and their reliance on large ``internal'' models and scaffolds makes them expensive to run, difficult to reproduce, and hard to study or improve upon. This raises a central question: \textit{can small, open models also be trained to achieve competitive reasoning performance on difficult Olympiad-level math?} In this paper, we answer this question by building QED-Nano, a 4B model post-trained for Olympiad-level proofs. Our training recipe has three stages: (1) supervised fine-tuning to imbue good proof-writing styles by distilling from DeepSeek-Math-V2, (2) reinforcement learning (RL) with rubric-based rewards, and (3) expanding RL with a reasoning cache, which decomposes long proofs into iterative summarize-and-refine cycles and enables stronger test-time reasoning. QED-Nano surpasses the proof-generation performance of much larger open models, including Nomos-1 and GPT-OSS-120B, and approaches the performance of proprietary models like Gemini 3 Pro, at a fraction of the inference cost. To support further research on open mathematical reasoning, we release the full QED-Nano pipeline, including the \href{https://huggingface.co/lm-provers/QED-Nano}{QED-Nano} and \href{https://huggingface.co/lm-provers/QED-Nano-SFT}{QED-Nano-SFT} models, the \href{https://huggingface.co/datasets/lm-provers/FineProofs-SFT}{FineProofs-SFT} and \href{https://huggingface.co/datasets/lm-provers/FineProofs-RL}{FineProofs-RL} datasets, and the \href{https://github.com/CMU-AIRe/QED-Nano}{training and evaluation code}.

\end{abstract}

\vspace{-0.3cm}
\section{Introduction}
\vspace{-0.2cm}
\begin{wrapfigure}{r}{0.52\textwidth}
\centering
\vspace{-8mm}
\includegraphics[width=0.99\linewidth]{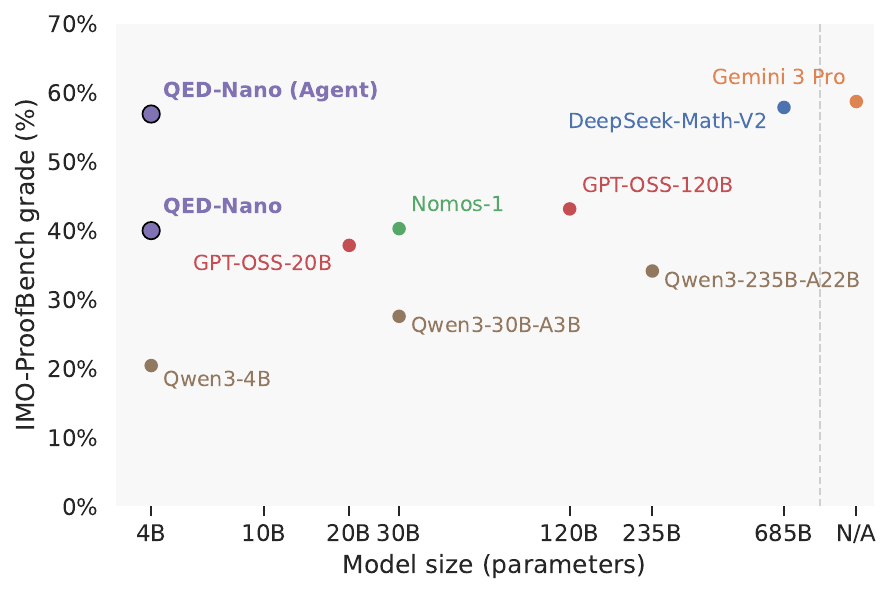}
\vspace{-6mm}
\caption{\footnotesize{\textbf{IMO-ProofBench \citep{imobench} grade as a function of model size.} QED-Nano outperforms open models of similar size and remains competitive with much larger systems.}}
\label{fig:model-size-vs-grade}
\vspace{-0.5cm}
\end{wrapfigure}

In the past year, large language models (LLMs) have made striking progress in mathematics, with some now solving complex proof-based problems at a level comparable to IMO gold medalists~\citep{imobench,shao2025deepseekmathv2}. However, the systems that achieve these results are either built using largely undisclosed training pipelines or rely on very large models with prohibitive training and inference costs. This makes them difficult to reproduce, study, or meaningfully extend, creating a widening gap between what is achievable in principle and what the broader research community can actually build and understand.

In this work, we ask whether small, open models can be post-trained to compete with much stronger proprietary systems on difficult Olympiad-level math proof problems. Our work is in the spirit of prior research showing that small models can be made surprisingly strong on challenging domains such as math problem solving and Olympiad-level coding, largely using combinations of supervised fine-tuning (SFT) on synthetic data~\citep{li2024common7blanguagemodels,penedo2025olympiccoder,moshkov2025aimo2winningsolutionbuilding} and basic instantiations of reinforcement learning (RL) with verifiable rewards~\citep{deepscaler2025}. However, Olympiad proof-writing is a substantially different setting: it requires generating long, globally coherent, and mathematically rigorous arguments, rather than arriving at a verifiable final answer or executable solution. We study what it takes to post-train a small open model to acquire this capability, despite the fact that such behavior lies far outside typical objectives or benchmarks (e.g., those based on verifiable answers) that were  used to guide training of our base model (i.e., Qwen3).

Concretely, we present an \emph{end-to-end post-training recipe for a 4B theorem-proving model}. Our model, QED-Nano, operates entirely in natural language, with no reliance on Lean or external tools. As shown in \cref{fig:model-size-vs-grade}, QED-Nano substantially outperforms open models of comparable size and remains competitive with much larger systems, including proprietary ones.
Our training pipeline has three components, resembling a standard post-training stack but with design choices tailored to Olympiad-level proof generation and effective use of test-time compute. First, we use SFT to give the model a basic ability to write mathematical proofs. Next, we apply rubric-based RL using an objective that \emph{explicitly} trains the model to improve its reasoning over the course of a long proof trajectory at test time~\citep{wu2026reasoningcache}. Finally, we develop test-time scaffolds that allow the model to fully exploit this learned iterative reasoning ability, improving performance for a given token budget.

\textbf{Main results.} Even \emph{without any test-time scaffold}, QED-Nano achieves a $40\%$ score on IMO-ProofBench \citep{imobench}, $45\%$ on ProofBench \citep{ma2025reliable}, and $68\%$ on IMO-AnswerBench \citep{imobench}, far better than any other 4B model. On average, these scores make QED-Nano outperform much larger open models such as Nomos-1 (30B) \citep{nomos2025} and Qwen3-235B-A22B-Thinking \citep{qwen3technicalreport}. More importantly, our main result shows that when allowed to reason for up to 2 million tokens per problem by pairing the model with a test-time scaffold, QED-Nano (Agent) achieves $57\%$ on IMO-ProofBench \citep{imobench}, $63\%$ on ProofBench, and $77\%$ on IMO-AnswerBench, attaining a strong cost-performance tradeoff on Olympiad-level problems (\cref{fig:model-size-vs-grade,tab:intro-model-comparison}). On IMO-ProofBench, this performance is close to Gemini 3 Pro, a strong proprietary model.

Beyond the results, we illustrate a broader principle: \emph{small models can be explicitly trained to adapt reliably at test time, even on highly challenging tasks.} Although we focus on Olympiad-style problems, the recipe is general and applies to other domains with rubric-based rewards. Since inference cost grows rapidly with model size, test-time adaptation is often more practical to scale with smaller models. Our results suggest that task-specialized small models can match or exceed much larger general-purpose systems, offering a path toward strong specialization without relying on extremely large and costly models.

\vspace{-0.2cm}
\section{Related Work} \label{sec:related_work}
\vspace{-0.2cm}

\textbf{LLMs for math reasoning.} RL is a key ingredient in improving the reasoning capabilities of LLMs~\citep{deepseekai2025deepseekr1,o1}, with approaches such as GRPO~\citep{shao2024deepseekmath} driving especially strong gains in mathematics. Recent reasoning models, including Qwen-3.5~\citep{qwen35}, GPT-5.4~\citep{openai2026gpt54}, and Step-3.5-Flash~\citep{huang2026step35flashopen}, now saturate challenging benchmarks such as AIME and HMMT~\citep{matharena}, and continue to make progress on harder benchmarks such as FrontierMath~\citep{frontiermath} and Humanity's Last Exam~\citep{hle}. Proprietary LLM systems are even beginning to support mathematical research directly, by helping with minor open problems~\citep{alphaevolvediscovery,aletheia,ghrist2024latticevaluedbottleneckduality,schmitt2025extremaldescendantintegralsmoduli,dobriban2025solving}. Yet open-source systems remain well behind these frontier capabilities, and the recipe required to close this gap is still poorly understood. Our work takes a step in this direction by showing that fairly small models can be post-trained to solve difficult Olympiad-level proof problems, yielding QED-Nano, one of the strongest open models for theorem proving.

\textbf{LLMs for natural language theorem proving.} Early reasoning-focused LLMs were trained primarily to optimize final-answer correctness~\citep{deepseekai2025deepseekr1,qwq32b}. However, later work has shown that such models often reach correct answers through shortcuts or shallow reasoning patterns~\citep{usamo,brainvbytes}, limiting their usefulness for advanced proof problems that require long, coherent, and rigorous arguments. 
This gap has motivated a growing body of work on theorem proving, including LLM judges for test-time reranking~\citep{opc,mahdavi2025scaling,ma2025reliable}, test-time scaffolds on  existing models~\citep{huang2025winninggold,imobench}, theorem proving in domain-specific languages such as Lean~\citep{chen2025seedprover15masteringundergraduatelevel, liu2026numinaleanagentopengeneralagentic}, and models trained specifically for natural language proof generation~\citep{shao2025deepseekmathv2,nomos2025}.

The last line of work is perhaps the most relevant to us. DeepSeekMath-V2~\citep{shao2025deepseekmathv2} introduces a 685B model tuned for proof writing, but its pipeline is expensive and difficult to reproduce due to the lack of training data or reward functions, and moreover requires human judgments at multiple stages. More comparable to our setting, Nomos-1~\citep{nomos2025} is a 30B model trained for the same task. While it is open-weight, it is not accompanied by a detailed training recipe. In contrast, QED-Nano is much smaller than both models, uses a fully open training pipeline, and outperforms Nomos-1 on theorem-proving benchmarks.

\textbf{RL techniques and test-time scaffolds.} Our work sits at the intersection of RL post-training, rubric-based reward design, and test-time scaling~\citep{snell2024scalingllmtesttimecompute}. Recent LLM post-training work has explored several policy-gradient methods, including GRPO~\citep{shao2024deepseekmath}, RLOO~\citep{ahmadian2024basicsrevisitingreinforcestyle}, and MaxRL~\citep{tajwar2026maximumlikelihoodreinforcementlearning}. In our setting, however, the specific RL algorithm is less important than rollout length, reward quality, and how training is coupled with test-time scaffolds. A growing body of work studies such scaffolds, including tree-of-thoughts~\citep{yao2023treethoughtsdeliberateproblem}, recursive aggregation methods~\citep{rsa},
and iterative summarization-based approaches~\citep{wu2026reasoningcache}.

In parallel, recent work has shown that structured rubric-based feedback can serve as a stable reward signal for RL beyond purely verifiable tasks~\citep{gunjal2025rubricsrewardsreinforcementlearning, he2025advancedifrubricbasedbenchmarkingreinforcement}. Finally, to achieve strong long-horizon performance, prior work often introduces curricula~\citep{setlur2025e3learningexploreenables,Polaris2025,deepscaler2025} or dense process rewards~\citep{qu2025optimizingtesttimecomputemeta} rather than training directly at long lengths. While we do not use curricula or process rewards, we do train with a test-time scaffold in the loop, since the proof lengths we target are too large for standard RL to optimize directly due to high-variance gradients~\citep{agarwal2021theory}.
\vspace{-0.2cm}
\section{Step I: Dataset Construction and Grading Schemes} \label{sec:dataset_construction}
\vspace{-0.2cm}

We start building our recipe by discussing how we curate our prompt sets and design our rubrics for RL training. To learn theorem-proving at the Olympiad level, we need proof questions that are both challenging and high-quality, with clear criteria for evaluating correctness and rigor. Therefore, we construct a high-quality corpus of around 5,000 problems that mirrors the structure and difficulty of IMO proofs as follows.

\begin{wrapfigure}{r}{0.5\textwidth}

\vspace{-0.6cm}
\begin{evalbox}{Grading Scheme Example}
\begingroup
\footnotesize
\setlength{\parindent}{0pt}
\begin{enumerate}[leftmargin=1em]
\item \textbf{Checkpoints (7 pts total)}
\begin{itemize}[leftmargin=1em,itemsep=-1pt]
\item \textbf{1 pts}: State or prove the inequality  [...]
\item \textbf{4 pts}: Boundedness of the sequence [...]
\item \textbf{2 pts}: Periodicity. [...]
\end{itemize}
\item \textbf{Zero-credit items}
\begin{itemize}[leftmargin=1em]
\item Claims that $d(n) < n$ [...] without [...]
\end{itemize}
\item \textbf{Deductions}
\begin{itemize}[leftmargin=1em]
\item \textbf{Cap at 5/7}: If the student proves $a_n$ is bounded but [...]
\end{itemize}
\end{enumerate}
\endgroup
\end{evalbox}
\vspace{-3mm}
\caption{\footnotesize{Example of a grading scheme.}}
\label{fig:grading_scheme_example}

\vspace{-0.6cm}
\end{wrapfigure}
\textbf{Data sources and filtering strategy.} We begin with two public datasets: \href{https://huggingface.co/datasets/AI-MO/aops}{AI-MO/aops}, which contains problems sourced from the Art of Problem Solving forums, and \href{https://huggingface.co/datasets/AI-MO/olympiads}{AI-MO/olympiads}, which aggregates official solutions from a wide range of national and international math competitions (\textit{e.g.}, IMO, USAMO, RMM, \textit{etc.}). While these sources provide coverage, they contain substantial noise, incomplete reasoning, formatting artifacts, and various other issues that preclude them from being seamlessly consumed in any post-training pipeline. We therefore apply a multi-stage filtering procedure to improve the data quality:

\vspace{-2mm}
\begin{enumerate}[itemsep=0em,leftmargin=1.2em]
\item We remove problems involving images, since our model operates purely in text.
\item We discard trivial or ill-posed entries, including problems where the answer appears directly in the statement,
solutions that are very short 
or purely computational, and materials drawn from easier contests such as AMC or routine exercises. To further enhance solution quality, we run an additional automated filtering pass using GPT-5-Nano (details in \cref{app:filtering_nano}). In particular, we prompt it to detect frequent issues observed in the \href{https://huggingface.co/datasets/AI-MO/aops}{AI-MO/aops} dataset, such as questionable problem statements, inconsistencies across proposed solutions, and reference proofs containing substantial logical gaps.
\item To avoid contamination with benchmarks, we exclude all problems from 2025 competitions and run a fuzzy string matching algorithm against IMO-ProofBench and ProofBench. The resulting dataset is a curated collection of Olympiad-style proof problems spanning geometry, number theory, algebra, and combinatorics.
\end{enumerate}
\vspace{-2mm}


\textbf{Grading schemes.} 
To provide accurate reward signals for training via RL, we construct detailed grading schemes for each problem. Our approach for constructing the grading scheme follows the framework introduced in \citet{ma2025reliable}, which uses Gemini 3 Pro with a custom prompt to generate rubrics that score model solutions from 0 to 7. Each rubric specifies: \textbf{(1)} detailed intermediate checkpoints corresponding to partial correctness; \textbf{(2)} common failure modes that warrant zero credit, and \textbf{(3)} specific points where additional deductions are necessary. An example is shown in \cref{fig:grading_scheme_example}. As a result, RL training receives dense, informative feedback instead of sparse success signals, encouraging gradual improvement in long-form reasoning rather than binary outcome optimization. In \cref{app:grading_schemes}, we provide more examples of the grading schemes we generate for our dataset.

\textbf{Problem difficulty annotations.} We annotate each problem with a difficulty estimate as determined by the average performance of our base model (Qwen3-4B-Thinking \citep{qwen3technicalreport}), computed over 128 parallel attempts, graded by GPT-OSS-20B \citep{gptoss}, and using the grading schemes mentioned above. We use these annotations to develop a difficulty-based learning curriculum during RL training.

\vspace{-0.3cm}
\section{Step II: Training Recipe}
\label{sec:training_recipe}
\vspace{-0.2cm}

In this section, we describe our post-training recipe for improving the proof-writing capabilities of small LLMs. In particular, we first discuss our core RL training approach, detailing the reward design and training curriculum (\cref{sec:core-rl}). To enable scaling performance with large test-time budgets, we then describe how we incorporate a scaffold based on iterative summarization and generation during training~\citep{wu2026reasoningcache} to optimize for behavior that benefits from large test-time budgets (\cref{sec:rc}). Finally, we discuss how we apply this recipe on top of a base model initialization that is capable of writing more capable proofs obtained via distillation from DeepSeek-Math-V2 \citep{shao2025deepseekmathv2} (\cref{sec:sft}).

\vspace{-0.2cm}
\subsection{Core Reinforcement Learning Approach} \label{sec:core-rl}
\vspace{-0.2cm}

We use GRPO \citep{shao2024deepseekmath} with rubric-based rewards as our core RL training approach. To adapt this algorithm to our setting, we first design a reward function based on a rigorous grading protocol and then run outcome-reward RL at a long response length.

\textbf{Grading protocol.} Designing a reliable reward signal for RL requires carefully balancing fidelity to human judgment with computational efficiency. A strong grader should produce scores that align closely with human evaluations while maintaining low latency, so that it remains practical for large-scale RL training\footnote{When using distributed asynchronous RL training infrastructure, grader latency is even more critical to tune as it alters the speed of sampler and trainer workers as well.}. 
We therefore conducted a series of experiments studying grader model choice, system instructions, and reasoning budget. Our initial pool included models with fewer than 120B active parameters from the GPT-OSS and Qwen3 families, but we found that Qwen3 models in this parameter range were unable to reliably evaluate all rubrics prescribed the grading scheme. Among GPT-OSS models, our preliminary results (see \cref{app:grading_protocol} for details) showed that GPT-OSS-20B with medium reasoning performs nearly as well as GPT-OSS-120B with high reasoning. We therefore use the smaller, lower-latency GPT-OSS-20B model as the judge during training. Finally, we use a prompt adapted from \citet{ma2025reliable}, augmented with the grading schemes.

\textbf{Outcome-reward RL at long lengths.}
Equipped with this grading scheme, we run RL to optimize the resulting outcome rewards. Two design choices remain when instantiating an RL run: the prompt set and the RL hyperparameters, in particular, the number of parallel rollouts per problem and the maximum response length~\citep{Cheng2026isocompute}. We construct a prompt set such that the base model's pass@1 scores follow a unimodal, heavy-tailed distribution, with a peak near difficult problems and a decreasing probability of sampling substantially easier ones, analogous to Polaris~\citep{Polaris2025}. Crucially, we remove all very easy problems on which the base model can attain a pass@1 score $\geq0.7$ and also remove problems where no rollouts succeed to improve training efficiency. With this prompt set, we now describe our workflow for setting various hyperparameters.


\begin{figure}[t]
\centering
\vspace{-0.3cm}
\includegraphics[width=0.47\linewidth]{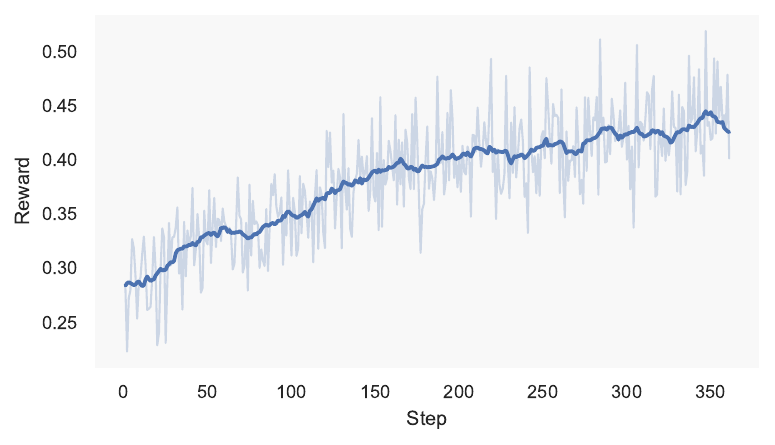}
\hfill
\includegraphics[width=0.47\linewidth]{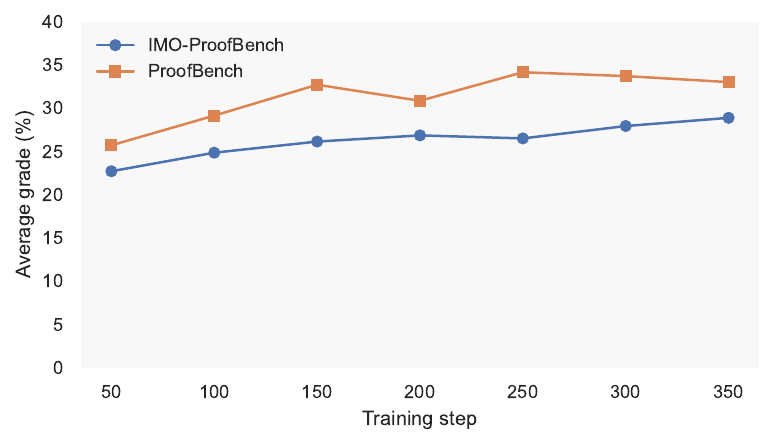}
\vspace{-0.3cm}
\caption{\footnotesize{RL training curves with rubric-based rewards (left) and corresponding evaluation metrics on IMO-ProofBench and ProofBench (right)}}
\label{fig:rl-training-dynamics}
\vspace{-0.5cm}
\end{figure}

\textbf{Base RL algorithm.} We use GRPO \citep{shao2024deepseekmath} as our base RL algorithm and build on PipelineRL \citep{piche2025pipelinerl} to implement an asynchronous, streaming variant of this algorithm. This implementation performs off-policy updates, with a maximum lag of 5 gradient steps between the current policy and the reference policy. We ablate several hyperparameters, including the number of parallel rollouts per problem, the entropy coefficient, and the KL divergence loss. We utilize an entropy coefficient of $0.0001$ through training and \emph{no} KL regularization. We find that a larger number of rollouts $n$ per problem improves performance when sufficient training epochs are run. Based on initial experiments with $n = 4, 8, 16$, we selected $n = 16$ because the fraction of problems on which no successful rollout is sampled is merely 2-3\% at $n=16$, which ensures effective utilization of training compute. Running this required 7 nodes for sampling on a batch of 64 problems (for a total of 1024 samples) and 4 nodes for the trainer.

We set the maximum response length to 50,000 tokens for RL training, since 95\% of responses from the base model terminate within this limit. As training progresses, however, we observe a noticeable increase in output length, consistent with observations from DeepSeek-R1 \citep{deepseekai2025deepseekr1} and others. A representative learning curve and corresponding evaluation scores are shown in \cref{fig:rl-training-dynamics}. We observe a noticeable increase in both the training and evaluation scores (on both of our benchmarks, IMO-ProofBench and ProofBench).



\begin{AIbox}{Summary: Core RL Recipe}
Rubric-based outcome rewards provide a stable signal for proof quality. We filter the prompt set to remove too easy and too hard questions. A moderate number of parallel rollouts ($n=16$) is sufficient for learning efficiently on our training data.
\end{AIbox}
\vspace{-0.2cm}
\subsection{RL for Continual Improvement at Test Time via Reasoning Cache} \label{sec:rc}
\vspace{-0.25cm}

\begin{wrapfigure}{r}{0.48\textwidth}
\centering
\vspace{-6mm}
\includegraphics[width=0.99\linewidth]{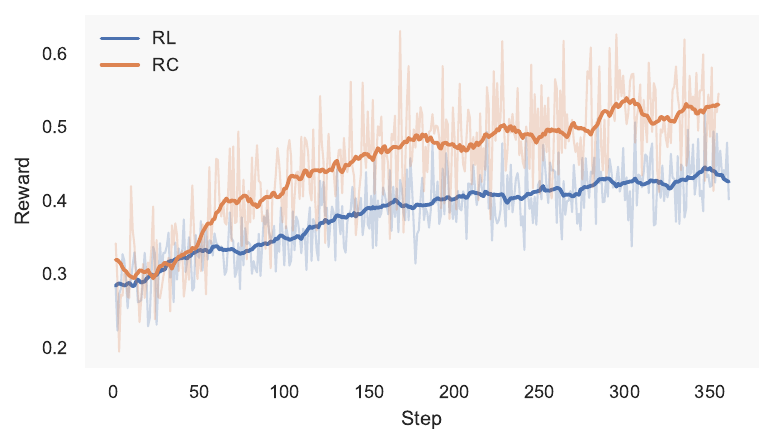}
\vspace{-0.7cm}
\caption{\footnotesize{\textbf{Training reward} as a function of optimization steps for standard RL and RL with RC.}}
\label{fig:rc-vs-rl-training}
\vspace{-0.3cm}
\end{wrapfigure}

Having established that RL improves both train- and test-time performance, the next step is to scale these gains further. For a small 4B model, increasing test-time compute offers a direct way to extract additional performance: if the model could think for longer and continue making \emph{progress} it could potentially solve harder problems better. A na\"ive approach would simply increase the maximum response length, and train at a matching output length, but this quickly becomes impractical due to high variance.

Instead, we introduce additional structure into the rollout generation process. Specifically, we adopt an iterative decoding procedure in which the model produces short reasoning segments that can be optimized with standard RL, while improving long-horizon performance. We implement this using \textbf{Reasoning Cache (RC)} framework \citep{wu2026reasoningcache}. RC decomposes reasoning into multi-step refinement cycles: at each iteration, the model generates a partial trace, summarizes its progress into a compact textual ``state representation'', and conditions the next rollout on both the original problem and this summary. The summary is then updated with the information produced in the current step. We train the model with RL to improve this summary-conditioned generation process. This structure enables effective exploration over reasoning horizons equivalent to hundreds of thousands of tokens while keeping training rollouts short and tractable as in the previous section.

\begin{wrapfigure}{r}{0.52\textwidth}
\centering
\vspace{-8mm}
\includegraphics[width=0.99\linewidth]{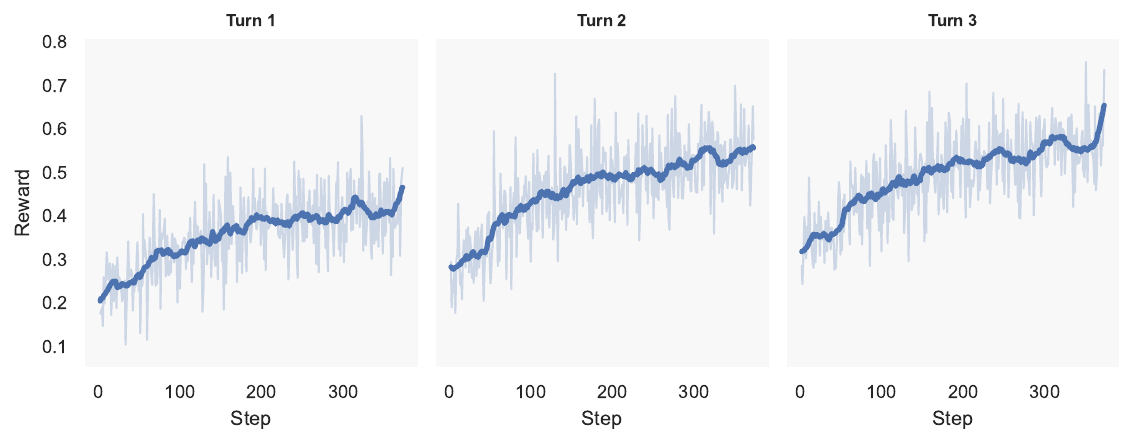}
\vspace{-8mm}
\caption{\footnotesize{Per-turn mean reward during RC training.}}
\label{fig:rc-turn-rewards}
\vspace{-0.3cm}
\end{wrapfigure}

We apply RL updates across these RC states, training the model to improve conditioned on the summary. Empirically, RC improves training stability, convergence speed, and performance compared to standard RL (\cref{fig:rc-vs-rl-training}) \emph{even within a single turn}. It also reduces the per decoding-turn response length, although this can be compensated for by running for more cycles. We observe an overall increasing trend in performance as more turns are performed during training (\cref{fig:rc-turn-rewards}).
While we use the same model for both reasoning and summarization at test-time, during training, we use a frozen snapshot of the Qwen3-4B-Instruct-2507 model as the summarizer for latency reasons.

\begin{wrapfigure}{r}{0.52\textwidth}
\centering
\vspace{-6mm}
\includegraphics[width=0.99\linewidth]{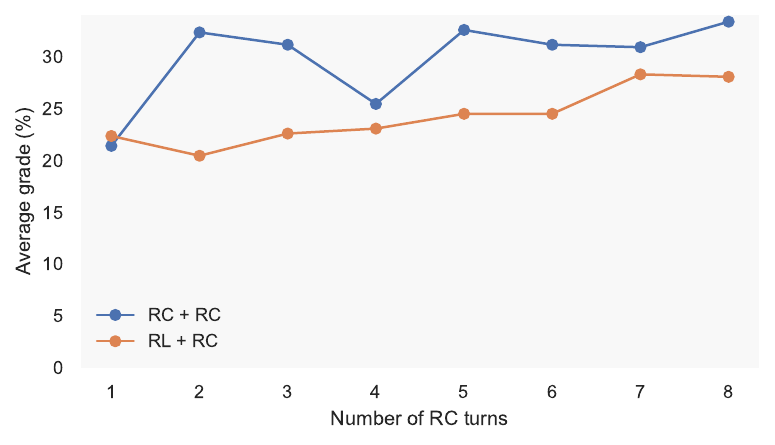}
\vspace{-6mm}
\caption{\footnotesize{Average grade on IMO-ProofBench versus RC turns.}}
\label{fig:rc-scaling-grade}
\vspace{-0.3cm}
\end{wrapfigure}

\textbf{Upon evaluation,} we find that both the RL-trained and RC-trained models achieve similar performance within a single decoding turn. However, the RC-trained checkpoint improves substantially more when run with the RC scaffold (\cref{fig:rc-scaling-grade}). In particular, the RC-trained model outperforms the RL-trained model at every turn, with the largest gap appearing within the first three turns, which matches the number of turns used during training.  These results suggest that RC-style training better prepares the model to benefit from test-time scaffolds, and therefore we adopt RC training in our final recipe.

\vspace{-0.2cm}
\subsection{Initializing Proof-Writing Capabilities via Supervised Fine-Tuning} \label{sec:sft}
\vspace{-0.2cm}

Despite the promising results from RL on top of the 4B model, we found that building coverage over certain proof-writing strategies with an SFT stage provides a better initialization for the RL run. Our SFT recipe fine-tunes the base model on problems paired with proof solutions generated by DeepSeek-Math-V2 \citep{shao2025deepseekmathv2}, a 685B model fine-tuned specifically for Olympiad math (with a complex training procedure that involves meta-verifiers). We distill this teacher’s reasoning traces into a dataset of $\approx$7.5k sampled responses suitable for fine-tuning our 4B base model. We describe this in detail below. Additional details and ablations are provided in \cref{app:sft_ablations}

\textbf{SFT dataset generation using DeepSeek-Math-V2.} We generate multiple solutions for problems in our curated dataset with a 128k-token context limit. We first filter generations to retain only structurally valid completions containing closed reasoning blocks and explicit proof sections. We \textbf{\emph{do not}} filter low-scoring or incorrect proofs because they might still provide useful information about proof writing. This process yields a dataset of \textbf{7,500 proofs} across 4,300 distinct olympiad-level problems that are drawn from the dataset in \cref{sec:dataset_construction}.

We fine-tuned our base 4B model on this dataset using a global batch size of 32 for five epochs. We applied a cosine learning rate schedule with a 10\% warmup and a peak value of $3 \times 10^{-5}$, which provided a low validation SFT loss on a held-out dataset at convergence. 

\textbf{Challenges and considerations with SFT.} We provide an extensive ablation study of SFT hyperparameters in \cref{app:sft_ablations}. While SFT gives a strong initialization, it also introduces substantial challenges, most notably \textbf{length explosion}, even under our best configuration. Although the training data caps sequences at 45k tokens, the fine-tuned model often produces outputs extending to hundreds of thousands of tokens, especially on incorrect proof attempts. Rather than learning structured long-form reasoning, the model often imitates the superficial form of long proofs~\citep{gudibande2023false}, repeating itself or meandering until the context window is exhausted. This is a natural consequence of offline training on data from a much stronger model, or more generally on targets that are difficult to fit, and it motivates a more experiential learning paradigm.

RL provides a natural mechanism for experiential learning, and in our experiments it also reduces response length. As discussed in \cref{sec:rc}, training with RC accelerates this drop. However, when initialized from an overly verbose SFT checkpoint, the early phase of RL remains heavily confounded by rollout truncation and high overflow rates, often averaging around 60\%. This weakens credit assignment and reduces the effectiveness of both standard RL and RC. We believe that reducing verbosity in the SFT checkpoint would allow RL to improve faster and by a larger margin, but we leave this to future work.

\begin{AIbox}{Summary: Final Training Recipe for QED-Nano}
QED-Nano is trained by combining SFT initialization with rubric-based RL in the Reasoning Cache (RC) framework. RC allows small models to better exploit large test-time compute by aligning training with an iterative inference-time scaffold. RL also helps correct SFT-induced pathologies, most notably length explosion.
\end{AIbox}
\vspace{-0.3cm}
\section{Results} \label{sec:results}
\vspace{-0.3cm}

In this section, we compare QED-Nano against state-of-the-art open and closed models, including multiple sizes from the Qwen3 family~\citep{qwen3technicalreport}, Nomos-1~\citep{nomos2025}, GPT-OSS~\citep{gptoss}, DeepSeek-Math-V2~\citep{shao2025deepseekmathv2}, and Gemini-3-Pro~\citep{deepmind2025gemini3pro}. We evaluate all models on three benchmarks: \textbf{(1)} ProofBench~\citep{ma2025reliable}, originally introduced for proof grading; \textbf{(2)} IMO-ProofBench; and \textbf{(3)} IMO-AnswerBench, which consists of IMO-level problems with verifiable final answers~\citep{imobench}. ProofBench and IMO-ProofBench evaluate proof-writing ability, while IMO-AnswerBench measures final-answer performance, even though QED-Nano is not trained on answer-only questions. Following prior work, we use Gemini-3-Pro as the judge with the prompts recommended by ProofBench and IMO-ProofBench.

\begin{table}[h]
\centering
\small
\caption{\footnotesize{\textbf{Comparison of QED-Nano (4B) with leading open- and closed-source models} on IMO-ProofBench, ProofBench, and IMO-AnswerBench benchmarks. We report the avg@3 grade (\%) for the proof-based benchmarks to account for their relatively small size compared to IMO-AnswerBench.}}
\vspace{-3mm}
\begin{tabular}{lccc}
\toprule
 Model & IMO-ProofBench & ProofBench & IMO-AnswerBench \\
\midrule
 Qwen3-4B-Thinking-2507 & 20.4 (2.6) & 19.5 (0.9) & 55.8 \\
 \textbf{QED-Nano (SFT initialization only)} & \textbf{39.5 (2.9)} & \textbf{33.3 (0.5)} & \textbf{57.5} \\
 \textbf{QED-Nano} & \textbf{40.0 (0.6)} & \textbf{44.9 (3.4)} & \textbf{67.5} \\
 \textbf{QED-Nano (+ RSA test-time scaffold)} & \textbf{56.9 (5.9)} & \textbf{62.6 (4.0)} & \textbf{76.5} \\
 \midrule
 Qwen3-30B-A3B-Thinking-2507 & 27.6 (1.0) & 26.1 (2.4) & 67.0 \\
 Qwen3-235B-A22B-Thinking-2507 & 34.1 (0.7) & 33.7 (1.1) & 70.5 \\
 Nomos-1 & 40.3 (3.5) & 28.3 (3.9) & 49.0 \\
 GPT-OSS-20B & 38.3 (1.2) & 38.4 (3.9) & 61.5 \\
 GPT-OSS-120B & 43.1 (3.2) & 47.5 (1.7) & 70.5 \\
 \midrule
 DeepSeek-Math-V2 (685B) & 57.9 (2.0) & 60.6 (0.1) & 75.8 \\
 Gemini 3 Pro (proprietary) & 58.7 (2.9) & 66.7 (3.1) & 83.2 \\
\bottomrule
\end{tabular}
\label{tab:intro-model-comparison}
\end{table}

\textbf{Summary of main results for the trained model.} Note in \cref{tab:intro-model-comparison}, QED-Nano performs competitively with much larger models: it performs close to GPT-OSS-120B, and it outperforms Nomos-1 despite being roughly $8\times$ smaller. The gains over QED-Nano-SFT are most pronounced on ProofBench and IMO-AnswerBench, where QED-Nano improves by about 10\% in accuracy. Finally, when evaluated with the RSA scaffold~\citep{rsa}, QED-Nano surpasses GPT-OSS-120B and approaches the performance of DeepSeek-Math-V2 despite being smaller. We discuss some of our results in detail below.

\textbf{Result \#1: Scaling test-time compute with QED-Nano.} Because our RL training procedure optimizes summary-conditioned generation without enforcing a fixed summary format, we expect the trained model to benefit from a broad class of inference-time scaffolds. These include methods that condition future generations on summaries of multiple past attempts, such as RSA, as well as methods that use detailed verification logs from previous generations to guide subsequent attempts, such as the DeepSeek Math scaffold. Indeed, we find that the trained model is able to verify and use information contained in these summaries. Motivated by this, we evaluate several alternative test-time scaffolds for further scaling QED-Nano's performance.

Specifically, we consider the following scaffolds: \textbf{(1)} Reasoning Cache (RC)~\citep{wu2026reasoningcache}, which is also used during training; \textbf{(2)} Self-Check~\citep{huang2025winninggold}, which does XYZ; \textbf{(3)} the scaffold used by Nomos~\citep{nomos2025}; \textbf{(4)} RSA~\citep{rsa}, which summarizes several parallel rollouts; and \textbf{(5)} the DeepSeekMath scaffold (DSM)~\citep{shao2025deepseekmathv2}, which interleaves generation and verification. Additional details on how these scaffolds are implemented are provided in \cref{app:scaffolds}.

In preliminary experiments on RL checkpoints trained on top of the base model, we evaluated several test-time scaffolds. DSM, RSA, and RC consistently gave the strongest gains, while other scaffolds helped only modestly. Among these, RC does not use parallel sampling at each turn, which makes it convenient for training but less effective for maximizing test-time performance. By contrast, DSM and RSA can scale parallel compute more effectively. We initially chose DSM because our SFT initialization used traces from DeepSeekMath-v2, making it a natural fit. However, we found that DSM is less suitable for answer-based evaluation, since it is designed specifically for proof problems (see \cref{app:scaffolds}). We therefore adopt RSA in our main experiments. Our main results highlight its effectiveness: although RSA uses substantially more tokens than DSM, it also delivers stronger performance.

\begin{table}[h]
\centering
\caption{\footnotesize{\textbf{Comparing test-time scaffolds with QED-Nano} on IMO-ProofBench, illustrating a trade-off between token usage and performance. In geeneral, we observe that the RSA and DeepSeekMath scaffolds are the most performant, with RSA needing more tokens but attaining the best performance.}}
\vspace{-1mm}
\label{tab:scaffold-comparison}
\small
\begin{tabular}{lccc}
\toprule
\textbf{Scaffold} & \textbf{Avg. grade} (\%) & \textbf{Avg. tokens} & \textbf{Tokens vs. single turn} \\
\midrule
Single Turn & 40.0 & 93,690 & 1.00$\times$ \\
Reasoning Cache (training) & 44.0 & 237,379 & 2.53$\times$ \\
DeepSeek Math & 54.0 & 1,605,879 & 17.14$\times$ \\
RSA & 56.9 & 2,045,764 & 21.84$\times$ \\
\bottomrule
\end{tabular}
\vspace{-0.3cm}
\end{table}

\textbf{Quantitative evaluation.} \Cref{tab:scaffold-comparison} summarizes the performance of different scaffolds applied to QED-Nano.
RC yields the smallest performance gain within 3 turns, but it also requires only about twice the number of tokens. This means it is suitable for training but not for attaining best performance at test time.
In contrast, RSA and DSM provide substantial improvements of $17\%$ and $14\%$, respectively.
However, they are substantially more expensive: RSA costs about 20 times as much as the base model, and DSM about 16 times as much. Overall, DSM provides a reasonable tradeoff for performance within a given token budget on proof problems, while RSA results in the best performance on both proof and final-answer problems. Therefore, we use RSA to report results in \cref{tab:intro-model-comparison}.

\begin{AIbox}{Results: QED-Nano attains strongest performance among open-source models}
\begin{itemize}[itemsep=2pt]
 \setlength{\leftskip}{-25pt}
    \item QED-Nano benefits from a broad range of test-time scaffolds, showing that training generalizes beyond any single decoding strategy (RC).
    \item Scaffolds that combine parallel and sequential computation, such as DSM and RSA, deliver the largest gains, though at substantially higher token cost.
    \item Together with the RSA scaffold, QED-Nano matches or exceeds the performance of the best open-source model (DeepSeekMath-v2) and is close to Gemini 3 Pro.
\end{itemize}
\end{AIbox}
\textbf{Result \#2: Qualitative Analysis of Solutions from QED-Nano.}
We manually examined a subset of QED-Nano’s generated proofs to assess two additional dimensions beyond benchmark scores: \textbf{(1)} whether the model attempts to reward-hack the LLM-based grader, and \textbf{(2)} the intrinsic quality of the proofs themselves. An experienced evaluator from our team, with substantial experience assessing LLM-generated proofs, reviewed a sample of solutions and compared their judgments with those of the automated grader. Overall, QED-Nano tends to produce proofs that are more structured, explicit, and easier to follow than the other small open models we compare against, while QED-Nano + scaffold (DSM, in this case) mainly improves by correcting substantial but fixable mistakes without noticeably changing the proof style. Detailed annotations are provided in \cref{app:graded_examples}, and \cref{tab:qualitative_writing_styles} summarizes the qualitative differences across the four systems.

We also paid particular attention to the possibility of reward hacking. We found no clear evidence of reward hacking against the Gemini 3 Pro grader. The human evaluator agreed with the automated grader on most problems, although the LLM grader was occasionally slightly more generous. Only one problem showed a substantial score discrepancy: the QED-Nano agent's solution to IMO 2025 Q2. Our human evaluator judged its computation-heavy approach incorrect, citing an algebraic error and too many gaps for a fully rigorous proof. We do not view this as deliberate reward hacking, since other models often attempted similar approaches on the same problem. See \cref{app:graded_examples} for more extensive details.
\begin{table}[h]
\caption{\footnotesize{Qualitative summary of the proof-writing style of the models.}}
\vspace{-3mm}
\label{tab:qualitative_writing_styles}
\small
\begin{tabular}{p{0.2\linewidth} p{0.75\linewidth}}
\toprule
\textbf{Model} & \textbf{Summary of proof style by human annotator} \\
\midrule
\textbf{QED-Nano} &
QED-Nano typically produces proofs that are well structured, explicit, and easy to follow. Its main stylistic weakness is a strong preference for computation-heavy arguments for geometry problems. \\

\textbf{QED-Nano ~~~~~~~~~~~~~~~~~+ scaffold} &
The scaffold version is stylistically very similar to the standalone model. The model mainly seems to improve over QED-nano on questions where QED-nano made significant, but fixable mistakes. \\

\textbf{Qwen-3-4B-Think} &
The base model often writes proofs that frequently contain obvious flaws such as circular reasoning, unjustified simplifications, and other errors that are clearly related to its overoptimization for final answer performance. \\

\textbf{Nomos-1} &
Nomos-1 has a much more concise, direct style. This can read elegantly but also makes hidden mistakes harder to detect. \\
\bottomrule
\end{tabular}
\vspace{-0.6cm}
\end{table}

\vspace{-0.3cm}
\section{Discussion, Conclusion, and Perspectives on Future Work}
\vspace{-0.3cm}

In this paper, we introduce QED-Nano to demonstrate that Olympiad-level problem solving is not reserved for frontier-scale models with 100B+ parameters. With a post-training recipe that first instills proof-writing ability via SFT and then applies RL to explicitly optimize long-horizon improvement through a scaffold, a 4B model can produce substantially stronger proofs than its base initialization and compete with much larger open models when paired with test-time compute. On IMO-ProofBench, our QED-Nano (Agent) closes much of the gap to Gemini 3 Pro while being at least \textbf{3x cheaper to run and requiring significantly lower training costs.}

Going forward, several avenues could further improve QED-Nano. The most immediate is improving the synergy between SFT and RL. In particular, mitigating the length explosion introduced by SFT would likely amplify the gains from RL by speeding up credit assignment. A second short-term direction is to refine the grader, for example by gradually tightening rubric penalties during training or using stricter rewards to encourage increasingly rigorous and polished proofs. Finally, incorporating hints or guidance during training~\citep{qu2025pope}, such as conditioning on plans from oracle solutions or the grading scheme, may help the model tackle harder problems and further scale RL.

Beyond these directions, more fundamental questions remain. One is how to imbue LLMs with the ability to synthesize genuinely novel ideas or “aha” insights on the hardest problems. Like most LLMs, QED-Nano tends to rely on computation-heavy approaches rather than identifying elegant structural insights early. This likely reflects the style of reasoning that current RL procedures optimize for. Designing scalable training paradigms that encourage broader exploration of reasoning strategies, rather than refinement of a single computational path, is therefore an important challenge to study in the future.




\section*{Author Contributions}

This is a team effort with members from CMU, Hugging Face, ETH Zurich, and Numina.

Our team members (in alphabetical order) are as follows:

\begin{itemize}
\item CMU: Aviral Kumar, Yuxiao Qu, Amrith Setlur, Ian Wu
\item Hugging Face: Edward Beeching, Lewis Tunstall
\item ETH Zurich: Jasper Dekoninck
\item Project Numina: Jia Li
\end{itemize}

All members contributed to the project substantially. Specifically:

\begin{itemize}
\item Yuxiao Qu developed the initial version of the grader and verifier-based RL approach, built the grading-scheme pipeline, curated and processed the proof datasets, and implemented the initial reasoning-cache and multi-turn training loops that started us in this direction. With Amrith Setlur and Lewis Tunstall, he ran a number of ablations that informed the final RL runs.
\end{itemize}

\begin{itemize}
\item Amrith Setlur adapted the PipelineRL infrastructure for the verifier-based RL approach, optimized RL configurations for stability and scale, implemented the asynchronous and streaming reasoning-cache RL training infrastructure, and proposed several ablations and algorithmic strategies for the training runs. With Yuxiao Qu and Lewis Tunstall, he ran a number of ablations that informed the final RL runs.
\end{itemize}

\begin{itemize}
\item Ian Wu, as primary author of the Reasoning Cache method, provided core technical guidance on RC experimentation, evaluation, and training pipelines, which shaped how it was utilized throughout the project.
\end{itemize}

\begin{itemize}
\item Edward Beeching led several evaluations, developed the synthetic data generation pipeline for DeepSeek-Math-V2, and ran the SFT ablations to analyse model length-control behavior and training dynamics.
\end{itemize}

\begin{itemize}
\item Lewis Tunstall led the large-scale RL infrastructure efforts, benchmarking and stabilizing multiple RL frameworks at the start of the project, optimized inference throughput, and ran the largest training and evaluation experiments. With Amrith Setlur and Yuxiao Qu, he ran a number of ablations that informed the final RL runs. He advised several other aspects of the project.
\end{itemize}

\begin{itemize}
\item Jasper Dekoninck led the benchmark design and ensured rigorous evaluations, benchmarked several test-time agent scaffolds and developed our final scaffold, built the IMO-ProofBench and ProofBench splits, filtered training datasets and created the grading schemes for them, designed the benchmarks for the RL grader, and led extensive model-based and human evaluations to ensure robustness and correlation with human proof quality.
\end{itemize}

\begin{itemize}
\item Jia Li curated and expanded high-quality AoPS and Olympiad datasets, developed grading-scheme generation workflows, and explored scalable problem synthesis and verification strategies in the project initially.
\end{itemize}

\begin{itemize}
\item Aviral Kumar advised the overall project and contributed to the ideas behind long-horizon training, curriculum design, reward formulation and some ideas on data construction.
\end{itemize}

\section*{Acknowledgements}

We thank Leandro von Werra, Andres Marafioti, Thibaud Frere, Graham Neubig, Sewon Min, Wenjie Ma, and Katerina Fragkiadaki for helpful discussions and feedback. AS, YQ, IW, and AK thank the FLAME center at CMU, the DeltaAI cluster, and the NAIRR program for providing GPU resources that supported a part of the experimental iteration. We thank Google Cloud for Gemini 3 Pro API credits. AS and AK thank the Laude Institute Slingshots program for support and feedback, and Braden Hancock and Andy Konwinski at the Laude Institute for discussions and feedback. EB and LT thank Hugo Larcher and Mathieu Morlon for keeping the GPUs running hot on the Hugging Face cluster. JD used compute from the Swiss AI Initiative supported by a grant from the Swiss National Supercomputing Centre (CSCS) under project ID a155 on Alps.

\bibliographystyle{colm2026_conference}
\bibliography{references}
\appendix
\crefalias{section}{appendix}
\Crefname{appendix}{Appendix}{Appendices}
\newpage

\section{Grading Protocol}\label{app:grading_protocol}
To determine which grader model for training, we conduct a series of experiments.

\textbf{Grader evaluation benchmarks.} We construct two benchmarks to evaluate our grader design. First, we aggregate all human annotations from the proof-based portion of MathArena \citep{matharena}, comprising 438 solutions across 22 problems. Second, to obtain a benchmark more representative of our training-time prompt distribution that we will query the grader on, we randomly sample 60 problems from our training corpus. For each problem, we generate four candidate solutions from our base 4B model and Qwen3-30B-Think \citep{qwen3technicalreport}. We grade these solutions using Gemini 3 Pro, instructed with a prompt adapted from the ProofBench paper \citep{ma2025reliable}, which we found to yield evaluations consistent with human judgment. We therefore treat Gemini 3 Pro's grades as the ground-truth reference in this benchmark.

\textbf{Grader evaluation metric.} Both benchmarks contain multiple solutions per problem, enabling calibrated comparisons through a problem-normalized \textit{advantage score}. For each problem $p_i$ and solution $y_j^i$ to problem $p_i$, we compute the unnormalized advantage $A_{i,j} = r_{i,j} - \overline{r_i}$, where $r_{i,j}$ is the grader-assigned reward to solution $y_j^i$, and $\overline{r_i}$ is the mean reward across all solutions to problem $p_i$. Grader accuracy is measured as the mean absolute difference between the candidate grader's advantages and the reference advantages. This formulation removes sensitivity to constant or benign shifts between graders, which is important because such shifts do not affect RL training with several parallel rollouts used in GRPO.

\textbf{Grader model and prompt.} Using the metric above, we evaluate five grader prompts drawn from prior work emphasizing different evaluation ideologies \citep{opc,mahdavi2025scaling,huang2025winninggold,ma2025reliable} (\cref{tab:matharena-grader-results}). On the MathArena subset, GPT-OSS-20B with medium reasoning performs best when paired with the strict ProofBench \citep{ma2025reliable} prompt, which emphasizes strict adherence to the rubric and rejects solutions that deviate from it. All evaluated prompts are shown in \cref{app:prompts}.

\begin{table}[ht]
\centering
\small
\caption{Results on the MathArena grading benchmark. Lower is better.}
\vspace{-3mm}
\begin{tabular}{lccccc}
\toprule
 Model & Simple & OPC & ProofBench & ProofBench Strict & GIMO \\
\midrule
 GPT-OSS-20B-medium & 1.56 & 1.57 & 1.43 & \textbf{1.21} & 1.36 \\
\bottomrule
\end{tabular}
\label{tab:matharena-grader-results}
\end{table}

We then compare the choice of grader models and evaluate whether including a reference proof alongside the marking scheme improves performance (\cref{tab:indistribution-grader-results}). We conduct this experiment on the in-distribution grading benchmark as it is more representative of scenarios that the grader will encounter during training. We observe that the performance differences between models are minimal. GPT-OSS-20B with medium reasoning performs on par with the alternatives while being significantly cheaper and faster, so we adopt it as our grader for training. Including a reference solution slightly degrades performance, so we exclude it from the final grader configuration.

\begin{table}[ht]
\centering
\small
\caption{Results on our in-distribution grading benchmark. Lower is better.}
\vspace{-3mm}
\begin{tabular}{lcc}
\toprule
 Model & ProofBench Strict & ProofBench Strict (with ref) \\
\midrule
 GPT-OSS-20B-medium & 1.19 & 1.26 \\
 GPT-OSS-20B-high & 1.17 & 1.19 \\
 GPT-OSS-120B-medium & 1.16 & 1.24 \\
\bottomrule
\end{tabular}
\label{tab:indistribution-grader-results}
\end{table}
\newpage
\section{Test-Time Scaffolds} \label{app:scaffolds}
In this section, we discuss each of the test-time scaffolds that we considered for further scaling performance of QED-Nano.

\begin{table}[t]
\centering
\caption{Effect of scaling the number of turns for the reasoning cache on ProofBench.}
\label{tab:rc-turns}
\small
\begin{tabular}{ccc}
\toprule
Number of turns & Avg. grade (\%) & Avg. tokens \\
\midrule
3 & 40.0 & 261,892  \\
5 & 42.4 & 434,367  \\
7 & 54.0 & 578,928  \\
\bottomrule
\end{tabular}

\end{table}

\begin{itemize}
\item \textbf{Reasoning-Cache (RC)} \citep{wu2026reasoningcache}: The decoding algorithm that we used for RL post-training, which iteratively summarizes the current attempt and conditions subsequent response generation on it. This approach does not utilize parallel sampling directly and we found that using three turns of summarize-then-refine gave the best tradeoff between performance and test-time compute (Table~\ref{tab:rc-turns}).
\item \textbf{Self-Check} \citep{huang2025winninggold}: A simple generate-verify-improve loop that ends when the verifier cannot find any flaws in the solution anymore.
\item \textbf{Nomos} \citep{nomos2025}: This scaffold first generates $n$ solutions and verifies each of them once. It then filters the solutions to only keep the $k$ best ones, after which it is run through a single “consolidation” stage, where the model is presented with all $k$ solutions and asked which group of solutions is most likely to be correct. Among that group, the agent runs a simple knockout tournament with an LLM-judge to select the best one.
\item \textbf{RSA} \citep{rsa}: This scaffold first generates $n$ solutions. In each subsequent stage, it then generates $n$ new solutions by randomly conditioning each new solution on $k$ existing ones. After several iterations, a random proof is selected. We make one minor improvement over the original design: instead of selecting an arbitrary proof, we run a knockout tournament with an LLM-judge on the solutions from the last stage.
\item \textbf{DeepSeek Math} \citep{shao2025deepseekmathv2} (DSM): This scaffold first generates $n$ solutions, each of which is self-evaluated $n$ times. Solutions are sorted by their average self-evaluated score, and the top $n$ solutions are improved by presenting the model with the solution and some of the feedback generated by the self-evaluation stage. These new solutions are added to the solution pool. This process is iterated several times before the solution with the highest overall score across all iterations is returned.
\end{itemize}

\textbf{DSM on IMO-AnswerBench.} As noted in \cref{sec:results}, the DSM framework is specifically optimized for proof-based problems. Its prompts instruct models to produce rigorous proofs and to reject answers containing superfluous reasoning. By contrast, strong performance on final-answer benchmarks often comes from solutions that skip some details or make reasonable simplifying assumptions while still reaching the correct answer. Under DSM, such answers are rejected and replaced with ones that appear more rigorous but avoid these simplifications, which can even make them less accurate. Indeed, on IMO-AnswerBench, the DSM scaffold lowers QED-Nano's performance from 66.3\% to 58.3\%, despite incurring substantially higher cost. This effect is model-dependent: for GPT-OSS-20B, DSM improves performance from 61.5\% to 69.8\%.
\section{SFT Ablations} \label{app:sft_ablations}
We perform two additional ablations on our SFT experiments. 
\newpage
\begin{wrapfigure}[13]{r}{0.52\textwidth}
\centering
\vspace{-6mm}
\includegraphics[width=\linewidth]{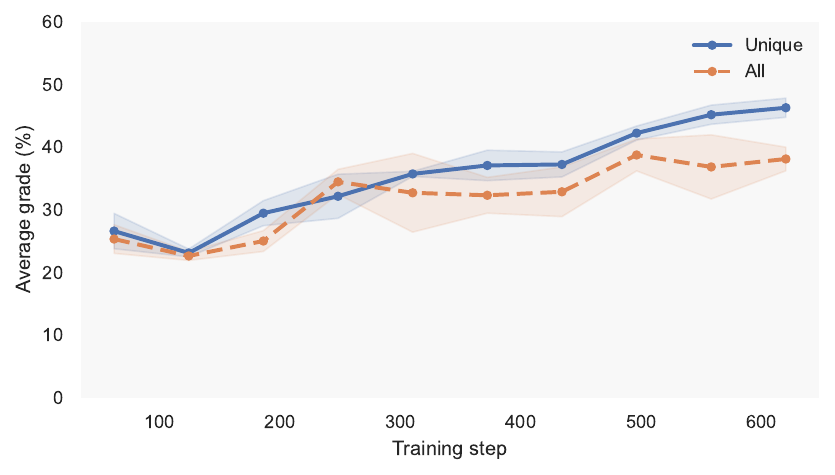}
\caption{Comparison between quantity and uniqueness in the SFT dataset.}
\label{fig:sft-data-ablation}
\end{wrapfigure}

\textbf{Data ablation: quantity vs. uniqueness.} We performed several ablations with different data mixtures; we highlight the comparison between training on the full corpus of 7,500 prompt-completion pairs versus a strictly filtered set of 4,300 correct solutions, where one solution is associated with a unique problem. As shown in \cref{fig:sft-data-ablation}, we find that training on only the unique problems achieves a higher final performance on IMO-ProofBench. The checkpoint at step 372 of this run was therefore used as an initialization for RL training.

\textbf{Model ablation.} We now perform an ablation over the base model and learning rate in our SFT experiments. We conducted these hyperparameter and model ablations to choose a suitable initialization for subsequent RL training. \Cref{tab:proofbench-best} reports the 620-step checkpoint for several learning rates and three different Qwen3-4B base models, Qwen3-4B Base, Qwen3-4B Instruct-2507, and Qwen3-4B Thinking-2507. Scores are shown as mean $\pm$ their standard deviation over 75 graded ProofBench problems. The base sweep remains weak across the tested learning rates, the instruct sweep improves substantially, and the thinking sweep gives the strongest overall ProofBench results. At the fixed 620-step checkpoint, the best performance comes from the Qwen3-4B Thinking-2507 sweep, where learning rates $2 \times 10^{-5}$ and $3 \times 10^{-5}$ tie on mean score. We use the $3 \times 10^{-5}$ model for further RL experiments.

\begin{table}[h]
\centering
\caption{ProofBench results for the Qwen3-4B SFT sweeps. Rows are grouped by base model and separated by horizontal rules. Within each group, each row corresponds to one learning rate and reports the mean $\pm$ SEM over 75 graded ProofBench problems at the 620-step checkpoint; boldface marks the best learning rate within each base-model group.}
\label{tab:proofbench-best}
\small
\setlength{\tabcolsep}{4pt}
\begin{tabular}{@{}lcc@{}}
\toprule
Base model & LR & ProofBench $\uparrow$ \\
\midrule
Qwen3-4B Base & $1 \times 10^{-6}$ & 0.00 $\pm$ 0.00 \\
 & $3 \times 10^{-6}$ & \textbf{0.29 $\pm$ 0.11} \\
 & $5 \times 10^{-6}$ & 0.07 $\pm$ 0.07 \\
 & $1 \times 10^{-5}$ & 0.21 $\pm$ 0.08 \\
 & $2 \times 10^{-5}$ & 0.09 $\pm$ 0.09 \\
 & $3 \times 10^{-5}$ & 0.19 $\pm$ 0.07 \\
 & $5 \times 10^{-5}$ & 0.08 $\pm$ 0.08 \\
 & $1 \times 10^{-4}$ & 0.12 $\pm$ 0.10 \\
\midrule
Qwen3-4B Instruct-2507 & $1 \times 10^{-6}$ & 0.03 $\pm$ 0.03 \\
 & $3 \times 10^{-6}$ & 0.77 $\pm$ 0.24 \\
 & $5 \times 10^{-6}$ & 1.95 $\pm$ 0.34 \\
 & $1 \times 10^{-5}$ & 1.37 $\pm$ 0.31 \\
 & $2 \times 10^{-5}$ & 2.09 $\pm$ 0.34 \\
 & $3 \times 10^{-5}$ & 1.72 $\pm$ 0.31 \\
 & $5 \times 10^{-5}$ & 1.79 $\pm$ 0.32 \\
 & $1 \times 10^{-4}$ & \textbf{2.11 $\pm$ 0.35} \\
\midrule
Qwen3-4B Thinking-2507 & $1 \times 10^{-6}$ & 2.03 $\pm$ 0.34 \\
 & $3 \times 10^{-6}$ & 1.65 $\pm$ 0.31 \\
 & $5 \times 10^{-6}$ & 2.28 $\pm$ 0.35 \\
 & $1 \times 10^{-5}$ & 2.52 $\pm$ 0.35 \\
 & $2 \times 10^{-5}$ & \textbf{2.85 $\pm$ 0.35} \\
 & $3 \times 10^{-5}$ & 2.85 $\pm$ 0.35 \\
 & $5 \times 10^{-5}$ & 2.36 $\pm$ 0.35 \\
 & $1 \times 10^{-4}$ & 1.93 $\pm$ 0.34 \\
\bottomrule
\end{tabular}

\end{table}

\section{Filtering with GPT-nano} \label{app:filtering_nano}

To further improve solution quality before training, we apply an additional automated filtering pass using GPT-5-Nano. Specifically, we prompt it to detect common issues observed in the \href{https://huggingface.co/datasets/AI-MO/aops}{AI-MO/aops} dataset, including questionable problem statements, inconsistencies among proposed solutions, and reference proofs with substantial logical gaps. Samples flagged as containing any of these issues are removed. We also initially asked the model to assess problem difficulty and whether a statement was purely computational, but we ultimately did not use these annotations because they were too noisy. Below, we provide the prompt used.
 
\begin{promptbox}{Filtering Prompt GPT-5-Nano}
\begingroup
\small
Your task is to classify this mathematical problem and solution among various categories. These problems and solutions were scraped automatically and therefore sometimes contain mistakes or inaccuracies.
Your feedback will be used to further filter the data to only keep high-quality samples.

You will need to output a json object with the following keys:
\begin{itemize}
\item \textbf{incomplete} (bool): If either the problem or solution is incomplete, you should mark this as true. This can occur if the solution only addresses part of the question, does not address the question at all, or if the question omits critical information without which certain quantities cannot be understood.
\item \textbf{difficulty} (int): an integer between 1 and 10 that indicates the difficulty of the problem. 1 indicates low difficulty questions on par with questions you would expect on the AMC. 4 are questions from national contests, 7 are the most difficult questions from national contests and the easiest questions from the IMO. 10 are the hardest questions from the IMO.
\item \textbf{gaps} (bool): This key should indicate whether the solution contains major gaps by considering several important aspects of the proof trivial or easy. Note, you should only mark this if the gaps are significant, and would form a big part of the proof if worked out.
\item \textbf{wrong} (bool): If the solution in any way indicates that the problem statement is ambiguous or wrong, you should mark this feature.
\item \textbf{disagreement} (bool): You will also see alternative solutions posted for the problem. If these alternative solutions/comments showcase disagreement on the accuracy of the selected solution, you should mark this. Note, it is is possible and allowed that there are several different approaches to the same problem. This field should only be marked if a user specifically points out a mistake or disagrees with the selected solution.
\item\textbf{computational} (bool): Is the main difficulty of the problem related to computing or evaluating a certain expression? We are interested in proof-based problems where the difficulty also lies in proving that the answer is X. Problems that are purely computational should therefore be excluded.
\end{itemize}
Format your reply as follows:

```json
\begin{quote}
\ttfamily
\small
\noindent
\{\\
\qquad "incomplete": \{\{true/false\}\}, \\
\qquad "difficulty": \{\{1/2/3/4/5/6/7/8/9/10\}\},\\
\qquad "gaps": \{\{true/false\}\},\\
\qquad "wrong": \{\{true/false\}\},\\
\qquad "disagreement": \{\{true/false\}\},\\
\qquad "computational": \{\{true/false\}\}\\
\}
\end{quote}
```

DO NOT forget the backticks (```json etc.) when you format your reply, they should be included.

\textbf{Problem}

\{problem\}

\textbf{Selected solution}

\{solution\}

\textbf{Problem Source}

\{path\}

\textbf{Alternative Solutions and Other Comments}

\{comments\}
\endgroup
\end{promptbox}
\section{Grader Prompts} \label{app:prompts}

We evaluate five grader prompts drawn from prior work emphasizing different evaluation ideologies (\cref{tab:matharena-grader-results}). All grader prompts are shown below.
\begin{promptbox}{ProofBench (Strict) Grader Prompt}
\begingroup
\small
\setlength{\parindent}{0pt}
\setlength{\parskip}{0.35em}
You are an \textbf{expert math proof grader}. You are judging the correctness of an LLM-generated proof for a math problem.

\textbf{Input}
\begin{itemize}
\item \textbf{Problem Statement}: A mathematical problem that the proof is attempting to solve.
\item \textbf{Marking Scheme}: A problem-specific grading rubric (0-7 scale) with checkpoints, zero-credit items, and deductions. You must follow this scheme when assigning points.
\item \textbf{Proof Solution}: The proof that you need to evaluate. This proof may contain errors, omissions, or unclear steps. The proof was generated by another language model.
\end{itemize}

\textbf{Task}

Analyze the proof carefully.

\textbf{Core principles (in order of precedence)}
\begin{enumerate}
\item \textbf{Mathematical validity} of the proof's reasoning and conclusion.
\item \textbf{Problem constraints} (e.g., unique required final value; forbidden tools if stated).
\item \textbf{Advisory mapping to the marking scheme} (checkpoints/deductions), allowing different orders and techniques.
\item \textbf{Reference solution} as an anchor for sufficiency, not exclusivity.
\end{enumerate}

\textbf{Alternative-approach policy}
\begin{itemize}
\item If the proof uses a different but valid method, \textbf{map its steps to equivalent rubric checkpoints} (same logical role) and award points accordingly.
\item Apply zero-credit items/deductions \textbf{only when the underlying issue actually occurs} in the given proof's approach; \textbf{auto-penalize} for omitting a rubric step.
\item Avoid double-counting mutually exclusive items; if two items solve the same logical gap, \textbf{award the larger only}.
\item If the final numeric/algebraic answer is wrong where uniqueness is required, award only partial credit justified by correct intermediate reasoning.
\end{itemize}

\textbf{Rigor and evidence}
\begin{itemize}
\item Award credit for intermediate claims \textbf{only if adequately justified} within the proof (not merely asserted).
\item If a step is plausible but under-justified, award \textbf{conservative partial credit} and note what is missing.
\end{itemize}

\textbf{What to produce}
\begin{itemize}
\item Identify logical errors, incorrect steps, or unclear reasoning.
\item Give a \textbf{score between 0 and 7} with a \textbf{detailed assessment}.
\item \textbf{Within the assessment text}, show clearly how the score was derived:
\begin{itemize}
\item Which rubric checkpoints were earned and the points you awarded.
\item Any zero-credit items or deductions you applied (and why).
\item How these add up to the final integer score in [0, 7].
\end{itemize}
\end{itemize}

\textbf{Output Format}

Respond with \textbf{only} well-formed XML using the structure below. Do not include any extra text or Markdown.

\textbf{Requirements}
\begin{itemize}
\item \texttt{\textless{}points\textgreater{}} must be an integer in [0, 7].
\item \texttt{\textless{}assessment\textgreater{}} must be a \textbf{detailed analysis} that explains your reasoning step-by-step and provides a clear \textbf{rationale for the score}. Reference specific claims/lines if present. Include the scoring breakdown \textbf{in prose} here (earned checkpoints or mapped equivalents, deductions, and subtotal to final score).
\item \texttt{\textless{}errors\textgreater{}} must be a list of specific issues (empty if score = 7).
\end{itemize}

\textbf{Example output}

\texttt{\textless{}points\textgreater{}0\textless{}/points\textgreater{}} \\
\texttt{\textless{}assessment\textgreater{}The proof shows a good understanding of the main idea, but has some unclear reasoning and minor mistakes...\textless{}/assessment\textgreater{}} \\
\texttt{\textless{}errors\textgreater{}} \\
\texttt{1. specific error 1,} \\
\texttt{2. specific error 2,} \\
\texttt{...} \\
\texttt{\textless{}/errors\textgreater{}}

\textbf{Input data}
\begin{itemize}
\item \textbf{Problem Statement}: \texttt{\{problem\}}
\item \textbf{Marking Scheme}: \texttt{\{marking\_scheme\}}
\item \textbf{Proof Solution}: \texttt{\{solution\}}
\end{itemize}
\endgroup
\end{promptbox}

\begin{promptbox}{ProofBench Grader Prompt}
\begingroup
\small
\setlength{\parindent}{0pt}
\setlength{\parskip}{0.35em}
You are an \textbf{expert math proof grader}. You are judging the correctness of an LLM-generated proof for a math problem.

\textbf{Input}
\begin{itemize}
\item \textbf{Problem Statement}: A mathematical problem that the proof is attempting to solve.
\item \textbf{Marking Scheme}: A problem-specific grading rubric (0-7 scale) with checkpoints, zero-credit items, and deductions. \textbf{Treat this scheme as advisory guidance, not a script.} Use it to anchor scoring, but \textbf{do not require} the proof to follow the same order, lemmas, or technique if its reasoning is mathematically sound.
\item \textbf{Proof Solution}: The proof that you need to evaluate. This proof may contain errors, omissions, or unclear steps. The proof was generated by another language model.
\end{itemize}

\textbf{Task}

Analyze the proof carefully.

\textbf{Core principles (in order of precedence)}
\begin{enumerate}
\item \textbf{Mathematical validity} of the proof's reasoning and conclusion.
\item \textbf{Problem constraints} (e.g., unique required final value; forbidden tools if stated).
\item \textbf{Advisory mapping to the marking scheme} (checkpoints/deductions), allowing different orders and techniques.
\end{enumerate}

\textbf{Alternative-approach policy}
\begin{itemize}
\item If the proof uses a different but valid method, \textbf{map its steps to equivalent rubric checkpoints} (same logical role) and award points accordingly.
\item \textbf{Do not penalize} solely for re-ordering steps, using different lemmas, or giving a correct shortcut, \textbf{unless} the problem forbids it.
\item Apply zero-credit items/deductions \textbf{only when the underlying issue actually occurs} in the given proof's approach; \textbf{do not auto-penalize} for omitting a rubric step that is unnecessary under the alternative method.
\item Avoid double-counting mutually exclusive items; if two items solve the same logical gap, \textbf{award the larger only}.
\item If the final numeric/algebraic answer is wrong where uniqueness is required, award only partial credit justified by correct intermediate reasoning.
\end{itemize}

\textbf{Rigor and evidence}
\begin{itemize}
\item Award credit for intermediate claims \textbf{only if adequately justified} within the proof (not merely asserted).
\item If a step is plausible but under-justified, award \textbf{conservative partial credit} and note what is missing.
\end{itemize}

\textbf{What to produce}
\begin{itemize}
\item Identify logical errors, incorrect steps, or unclear reasoning.
\item Give a \textbf{score between 0 and 7} with a \textbf{detailed assessment}.
\item \textbf{Within the assessment text}, show clearly how the score was derived:
\begin{itemize}
\item Which rubric checkpoints (or their mapped equivalents) were earned and the points you awarded.
\item Any zero-credit items or deductions you applied (and why).
\item How these add up to the final integer score in [0, 7].
\end{itemize}
\end{itemize}

\textbf{Output format}

Respond with \textbf{only} well-formed XML using the structure below. Do not include any extra text or Markdown.

\textbf{Requirements}
\begin{itemize}
\item \texttt{\textless{}points\textgreater{}} must be an integer in [0, 7].
\item \texttt{\textless{}assessment\textgreater{}} must be a \textbf{detailed analysis} that explains your reasoning step-by-step and provides a clear \textbf{rationale for the score}. Reference specific claims/lines if present. Include the scoring breakdown in prose here (earned checkpoints or mapped equivalents, deductions, and subtotal to final score).
\item \texttt{\textless{}errors\textgreater{}} must be a list of specific issues (empty if score = 7).
\end{itemize}

\textbf{Example output}

\texttt{\textless{}points\textgreater{}0\textless{}/points\textgreater{}} \\
\texttt{\textless{}assessment\textgreater{}The proof shows a good understanding of the main idea, but has some unclear reasoning and minor mistakes...\textless{}/assessment\textgreater{}} \\
\texttt{\textless{}errors\textgreater{}} \\
\texttt{1. specific error 1,} \\
\texttt{2. specific error 2,} \\
\texttt{...} \\
\texttt{\textless{}/errors\textgreater{}}

\textbf{Input data}
\begin{itemize}
\item \textbf{Problem Statement}: \texttt{\{problem\}}
\item \textbf{Marking Scheme}: \texttt{\{marking\_scheme\}}
\item \textbf{Proof Solution}: \texttt{\{solution\}}
\end{itemize}
\endgroup
\end{promptbox}

\begin{promptbox}{GIMO Grader Prompt}
\begingroup
\small
\setlength{\parindent}{0pt}
\setlength{\parskip}{0.35em}
You are an expert mathematician and a meticulous grader for an International Mathematical Olympiad (IMO) level exam.

Your primary task is to rigorously verify the provided mathematical solution using the provided marking scheme.

A solution that arrives at a correct final answer through flawed reasoning, educated guesses, or with gaps in its arguments must be flagged as incorrect or incomplete.

\textbf{Instructions}
\begin{enumerate}
\item \textbf{Core instructions}
\begin{itemize}
\item Your sole task is to find and report all issues in the provided solution. You must act as a \textbf{verifier}, not a solver. \textbf{Do not attempt to correct the errors or fill the gaps you find.}
\item You must perform a \textbf{step-by-step} check of the entire solution using the marking scheme. This analysis will be presented in a \textbf{Detailed Verification Log}, where you justify your assessment of each step: for correct steps, a brief justification suffices; for steps with errors or gaps, you must provide a detailed explanation.
\end{itemize}
\item \textbf{How to handle issues in the solution}

When you identify an issue in a step, you must first classify it into one of the following two categories and then follow the specified procedure.
\begin{itemize}
\item \textbf{Critical Error}: Any error that breaks the logical chain of the proof. This includes both \textbf{logical fallacies} (e.g., claiming that ``$A>B$, $C>D$ implies $A-C>B-D$'') and \textbf{factual errors} (e.g., a calculation error like ``$2+3=6$'').
\begin{itemize}
\item \textbf{Procedure}: Explain the specific error and state that it \textbf{invalidates the current line of reasoning}.
\end{itemize}
\item \textbf{Justification Gap}: A step where the conclusion may be correct, but the provided argument is incomplete, hand-wavy, or lacks sufficient rigor.
\begin{itemize}
\item \textbf{Procedure}: Explain the gap in the justification.
\item State that you will \textbf{assume the step's conclusion is true} for the sake of argument.
\item Then proceed to verify all subsequent steps to check if the remainder of the argument is sound.
\end{itemize}
\end{itemize}
\item \textbf{Output format}

Your response must be structured into three XML sections with tags \texttt{\textless{}summary\textgreater{}}, \texttt{\textless{}detailed\_verification\textgreater{}}, and \texttt{\textless{}points\textgreater{}}.
\begin{itemize}
\item \textbf{Summary}: Wrap this section within \texttt{\textless{}summary\textgreater{}...\textless{}/summary\textgreater{}} tags. This section must be at the very beginning of your response.
\begin{itemize}
\item \textbf{Final Verdict}: A single clear sentence declaring overall validity.
\item \textbf{List of Findings}: A bulleted list summarizing \textbf{every} issue, including:
\begin{itemize}
\item \textbf{Location}: A direct quote of the key phrase/equation.
\item \textbf{Issue}: A brief description and its classification (\textbf{Critical Error} or \textbf{Justification Gap}).
\end{itemize}
\end{itemize}
\item \textbf{Detailed Verification Log}: Wrap within \texttt{\textless{}detailed\_verification\textgreater{} ... \textless{}/detailed\_verification\textgreater{}} tags. Provide full step-by-step verification against the marking scheme, quoting the relevant solution text before analysis.
\item \textbf{Judgement}: Must be at the end of the response. For correct solutions, end exactly with \texttt{\textless{}points\textgreater{}INTEGER\textless{}/points\textgreater{}}. The score is the integer sum of points from marking-scheme steps.
\end{itemize}
\end{enumerate}

\textbf{Example of required summary format}

\texttt{\textless{}summary\textgreater{}} \\
\texttt{Final Verdict: The solution is invalid because it contains a Critical Error.} \\
\texttt{List of Findings:} \\
\texttt{Location: "By interchanging the limit and the integral, we get..."} \\
\texttt{Issue: Justification Gap - Missing justification for exchanging limit and integral.} \\
\texttt{Location: "From $A>B$ and $C>D$, it follows that $A-C>B-D$"} \\
\texttt{Issue: Critical Error - Logical fallacy.} \\
\texttt{\textless{}/summary\textgreater{}} \\
\texttt{\textless{}detailed\_verification\textgreater{}} \\
\texttt{...} \\
\texttt{\textless{}/detailed\_verification\textgreater{}} \\
\texttt{\textless{}points\textgreater{}0\textless{}/points\textgreater{}}

\textbf{Input data}
\begin{itemize}
\item \textbf{Problem Statement}: \texttt{\{problem\}}
\item \textbf{Marking Scheme}: \texttt{\{marking\_scheme\}}
\item \textbf{Proof Solution}: \texttt{\{solution\}}
\end{itemize}
\endgroup
\end{promptbox}

\begin{promptbox}{Simple Grader Prompt}
\begingroup
\small
\setlength{\parindent}{0pt}
\setlength{\parskip}{0.35em}
You are an \textbf{expert mathematics proof grader}. Your role is to evaluate the correctness, rigor, and completeness of a model-generated proof for a given problem using a provided \textbf{reference solution} and \textbf{grading schema}.

\textbf{Input components}
\begin{enumerate}
\item \textbf{Problem Statement}: the math problem to be solved.
\item \textbf{Marking Scheme (Schema)}: a JSON list of checkpoints (\texttt{title}, \texttt{desc}, \texttt{points}).
\item \textbf{Proof Solution}: the model-generated proof.
\end{enumerate}

\textbf{Your task}

Analyze the \textbf{Proof Solution} against the \textbf{Problem} and \textbf{Schema}, determine correspondence with rubric checkpoints, and assign an integer score \textbf{0-\{max\_points\}}.

\textbf{Output format}

Respond \textit{only} in XML:

\texttt{\textless{}assessment\textgreater{}DETAILED\_EVALUATION\_TEXT\textless{}/assessment\textgreater{}} \\
\texttt{\textless{}errors\textgreater{}} \\
\texttt{1. description of first issue,} \\
\texttt{2. description of second issue,} \\
\texttt{...} \\
\texttt{\textless{}/errors\textgreater{}} \\
\texttt{\textless{}points\textgreater{}INTEGER\textless{}/points\textgreater{}}

\textbf{Input data}
\begin{itemize}
\item \textbf{Problem Statement}: \texttt{\{problem\}}
\item \textbf{Reference Solution}: \texttt{\{reference\_solution\}}
\item \textbf{Marking Scheme}: \texttt{\{marking\_scheme\}}
\item \textbf{Proof Solution}: \texttt{\{solution\}}
\end{itemize}
\endgroup
\end{promptbox}

\begin{promptbox}{OPC Grader Prompt}
\begingroup
\small
\setlength{\parindent}{0pt}
\setlength{\parskip}{0.35em}
You are judging the correctness of an LLM-generated proof for a math problem.

\textbf{Input}
\begin{itemize}
\item \textbf{Problem Statement}: A mathematical problem that the proof is attempting to solve.
\item \textbf{Marking Scheme}: A marking scheme that rigorously specifies the steps in a correct proof to the problem. Any model should follow these steps. You should use this marking scheme to grade the solution.
\item \textbf{Proof Solution}: The proof that you need to evaluate. This proof may contain errors, omissions, or unclear steps. The proof was generated by another LLM.
\end{itemize}

\textbf{How the solution should be graded}

You should follow the marking scheme exactly. In particular, for each step in the marking scheme:
\begin{enumerate}
\item Analyze where this step was performed by the model and whether it occurs in the proof.
\item Analyze whether this step was performed sufficiently rigorously to award points. Any skipped steps or vague statements should be punished in your score.
\end{enumerate}

\textbf{Output format}

Give your analysis in the following format:

\texttt{\textless{}overall\textgreater{}} \\
\texttt{An overall analysis of the proof} \\
\texttt{\textless{}/overall\textgreater{}} \\
\texttt{\textless{}step\textgreater{}} \\
\texttt{For each step in the marking scheme, explain how many points you awarded for that step} \\
\texttt{\textless{}/step\textgreater{}} \\
\texttt{\textless{}points\textgreater{}INTEGER\textless{}/points\textgreater{}}

Here, \texttt{\textless{}points\textgreater{}} should be an integer summing up your scores for all separate steps in the marking scheme.

\textbf{Common mistakes and errors}

The proof you will evaluate is generated by an LLM. LLMs often bluff, hiding their mistakes in subtle areas that are difficult to detect.

Be very careful and strict, analyze and verify all statements made by the LLM. In particular:
\begin{itemize}
\item \textbf{Minor mistakes}: The following mistakes should be paid attention to, but usually only lose one or two points:
\begin{itemize}
\item Makes a small computational mistake that can be easily fixed.
\item Misses an edge case which can be easily proven/disproven.
\item Skips over a step that follows without much reasoning or manual work.
\end{itemize}
\item \textbf{Major mistakes}: The following mistakes are more problematic and usually cost a significant number of points, potentially invalidating the proof:
\begin{itemize}
\item Marks a step as trivial when it is not immediately obvious with little reasoning.
\item Omits algebra-heavy computational steps, regardless of whether methodology was outlined. Skipping short computations may be permitted.
\item Generalizes over a pattern without rigorously describing it or proving relevant properties.
\item Cites a non-existing or obscure theorem/source that cannot be immediately found online. Widely available theorems are allowed.
\end{itemize}
\end{itemize}

\textbf{Input data}
\begin{itemize}
\item \textbf{Problem Statement}: \texttt{\{problem\}}
\item \textbf{Marking Scheme}: \texttt{\{marking\_scheme\}}
\item \textbf{Proof Solution}: \texttt{\{solution\}}
\end{itemize}
\endgroup
\end{promptbox}

\section{Grading Scheme Examples} \label{app:grading_schemes}

This section contains the grading scheme examples shown in the blog post.

\subsection{Example 1}
\begin{promptbox}{Problem}
\begingroup
\small
\setlength{\parindent}{0pt}
Let $c$ be fixed natural number. Sequence $(a_n)$ is defined by: \newline
$a_1=1, a_{n+1}=d(a_n)+c$ for $n=1,2,...$. \newline
where $d(m)$ is number of divisors of $m$. Prove that there exist $k$ natural such that sequence $a_k,a_{k+1},...$ is periodic.
\endgroup
\end{promptbox}

\begin{evalbox}{Grading Scheme}
\begingroup
\small
\setlength{\parindent}{0pt}
\begin{enumerate}
\item \textbf{Checkpoints (7 pts total)}
\begin{itemize}
\item \textbf{1 pt}: State or prove the inequality $d(m) \leq \frac{m}{2} + 1$ (or a stronger bound such as $2\sqrt{m}$ for large $m$) to be used in the boundedness proof.
\item \textbf{4 pts}: Boundedness of the sequence $(a_n)$.
\begin{itemize}
\item \textbf{2 pts}: Combine the divisor bound with the recurrence to establish an inequality of the form $a_{n+1} \leq \frac{a_n}{2} + C$ (or equivalent logic showing $a_{n+1} < a_n$ for sufficiently large $a_n$).
\item \textbf{2 pts}: Conclude that the sequence is bounded (either globally bounded by a value like $2c+1$ using induction/contradiction, or eventually bounded via infinite descent).
\end{itemize}
\item \textbf{2 pts}: Periodicity.
\begin{itemize}
\item \textbf{1 pt}: Apply the Pigeonhole Principle to show that a value in the sequence must repeat.
\item \textbf{1 pt}: Conclude that repetition implies periodicity because the recurrence relation $a_{n+1} = d(a_n) + c$ is deterministic.
\end{itemize}
\item \textbf{Total (max 7)}
\end{itemize}
\item \textbf{Zero-credit items}
\begin{itemize}
\item Claims that $d(n) < n$ implies boundedness without a specific quantitative argument (since $a_{n+1} \approx a_n + c$ allows growth if $d(n) \approx n$).
\item Proving periodicity only for specific values of $c$.
\item Stating that the sequence is periodic because it is bounded, without proving boundedness.
\end{itemize}
\item \textbf{Deductions}
\begin{itemize}
\item \textbf{Cap at 5/7}: If the student proves $a_n$ is bounded but fails to explicitly mention the Pigeonhole Principle or finite states to deduce repetition.
\item \textbf{-1 point}: If the logic for boundedness relies on a bound like $d(n) \leq \sqrt{n}$ for all $n$ (which is false for small $n$), unless the argument is explicitly restricted to ``sufficiently large $n$''.
\item \textbf{No deduction}: For stating $d(m) \leq m/2 + 1$ without proof.
\item \textbf{No deduction}: For proving the sequence is eventually bounded rather than bounded for all $n$ (both are sufficient for the problem).
\end{itemize}
\end{enumerate}
\endgroup
\end{evalbox}

\subsection{Example 2}
\begin{promptbox}{Problem}
\begingroup
\small
\setlength{\parindent}{0pt}
Point $O$ is inside $\triangle ABC$. The feet of perpendicular from $O$ to $BC,CA,AB$ are $D,E,F$. Perpendiculars from $A$ and $B$ respectively to $EF$ and $FD$ meet at $P$. Let $H$ be the foot of perpendicular from $P$ to $AB$. Prove that $D,E,F,H$ are concyclic.
\endgroup
\end{promptbox}
\begin{evalbox}{Grading Scheme}
\begingroup
\small
\setlength{\parindent}{0pt}
\begin{enumerate}
\item \textbf{Checkpoints (7 pts total)}
\begin{itemize}
\item \textbf{Chain A: Main Isogonal Approach}
\item \textbf{Isogonality at Vertex A (3 pts)}
\begin{itemize}
\item Observe that $A, F, O, E$ are concyclic with diameter $AO$ (1 pt).
\item Prove that $AO$ and $AP$ are isogonal conjugates with respect to $\angle BAC$ (i.e., $\angle OAB = \angle PAC$), or equivalent angle chase establishing the direction of $AP$ (2 pts).
\item Note: Citing the ``isogonal lines lemma'' (altitude and circumradius are isogonal) for $\triangle AEF$ earns full marks for this checkpoint.
\end{itemize}
\item \textbf{Global Isogonality (1 pt)}
\begin{itemize}
\item Extend the logic to vertex $B$ (or state ``similarly'') to conclude that $P$ and $O$ are isogonal conjugates with respect to $\triangle ABC$.
\end{itemize}
\item \textbf{Concyclicity / Pedal Circle Theorem (3 pts)}
\begin{itemize}
\item Identify $D, E, F$ as the pedal triangle of $O$ and $H$ as a pedal point of $P$ (1 pt).
\item State/apply the theorem that pedal points of isogonal conjugates lie on a common circle to conclude $D, E, F, H$ are concyclic (2 pts).
\end{itemize}
\item \textbf{Chain B: Direct Angle Chasing (Alternative)}
\item \textbf{Characterizing P (4 pts)}
\begin{itemize}
\item Establish necessary angle properties of $P$ (equivalent to deriving isogonality) by calculating angles such as $\angle PAB$ in terms of angles related to $O$.
\end{itemize}
\item \textbf{Proving Concyclicity (3 pts)}
\begin{itemize}
\item Use the location of $P$ to prove $D, E, F, H$ are concyclic directly (e.g., via power of a point or proving $\angle DHE = \angle DFE$ adjusted for point order).
\end{itemize}
\item \textbf{Total (max 7)}
\end{itemize}
\item \textbf{Zero-credit items}
\begin{itemize}
\item Claiming $P$ is the isogonal conjugate of $O$ without any proof or derivation (conjecture).
\item Claiming $D, E, F$ lie on the nine-point circle or assuming $O$ is a specific center (e.g., circumcenter/orthocenter) without loss of generality.
\item Merely restating the definition of pedal points or concyclicity without progress.
\end{itemize}
\item \textbf{Deductions}
\begin{itemize}
\item \textbf{-1} for minor notation errors or mislabeling that does not affect the logical flow.
\item \textbf{-2} for misstating the pedal circle theorem (e.g., claiming the pedal circles of any two points coincide).
\item \textbf{Cap at 3/7} if the student assumes $P$ is the isogonal conjugate of $O$ without proof, even if the subsequent application of the pedal circle theorem is correct.
\end{itemize}
\end{enumerate}
\endgroup
\end{evalbox}

\subsection{Example 3}
\begin{promptbox}{Problem}
\begingroup
\small
\setlength{\parindent}{0pt}
An *up-right path* from $(a, b) \in \mathbb{R}^2$ to $(c, d) \in \mathbb{R}^2$ is a finite sequence $(x_1, y_z), \dots, (x_k, y_k)$ of points in $ \mathbb{R}^2 $ such that $(a, b)= (x_1, y_1), (c, d) = (x_k, y_k)$ , and for each $1 \le i < k$ we have that either $(x_{i+1}, y_{y+1}) = (x_i+1, y_i)$ or $(x_{i+1}, y_{i+1}) = (x_i, y_i + 1)$ . Two up-right paths are said to intersect if they share any point.

Find the number of pairs $(A, B)$ where $A$ is an up-right path from $(0, 0)$ to $(4, 4)$ , $B$ is an up-right path from $(2, 0)$ to $(6, 4)$ , and $A$ and $B$ do not intersect.
\endgroup
\end{promptbox}
\begin{evalbox}{Grading Scheme}
\begingroup
\small
\setlength{\parindent}{0pt}
\begin{enumerate}
\item \textbf{Checkpoints (7 pts total)}
\begin{itemize}
\item \textbf{1 pt}: Correctly identifies the total number of unrestricted path pairs as $\binom{8}{4}^2$ (or $70^2 = 4900$).
\item \textbf{2 pts}: Applies the Reflection Principle (or Lindstr\"om-Gessel-Viennot lemma) strategy. This involves establishing a correspondence between intersecting pairs of paths and pairs of paths with swapped endpoints.
\item Note: Award these points if the student describes the ``cut and swap tails at the first intersection'' bijection or correctly sets up the LGV determinant.
\item \textbf{1 pt}: Identifies the correct target set for the bijection: one path from $(0,0)$ to $(6,4)$ and one path from $(2,0)$ to $(4,4)$.
\item Note: Implicitly awarded if the student correctly writes the term $\binom{10}{4}\binom{6}{2}$.
\item \textbf{1 pt}: Justifies the validity of the subtraction. Specifically, notes that \textbf{every} pair of paths in the swapped configuration must intersect (guaranteeing the bijection covers the entire set), or cites the LGV lemma which accounts for this.
\item \textbf{1 pt}: Correctly evaluates the number of intersecting/swapped pairs as $\binom{10}{4}\binom{6}{2}$ (or $210 \times 15 = 3150$).
\item \textbf{1 pt}: Correct final answer: $1750$.
\item \textbf{Total (max 7)}
\end{itemize}
\item \textbf{Zero-credit items}
\begin{itemize}
\item Calculating $\binom{8}{4}^2$ (total pairs) but failing to make any progress on counting intersecting paths.
\item Subtracting an arbitrary term without justification derived from the path geometry or reflection principle.
\item Attempts to count intersections by summing over specific grid points without a valid summation method (this approach is generally intractable).
\end{itemize}
\item \textbf{Deductions}
\begin{itemize}
\item \textbf{-1}: Arithmetic errors in evaluating binomial coefficients or the final subtraction (e.g., $4900 - 3150 \neq 1750$).
\item \textbf{-1}: Misidentifying the endpoints for the swapped paths (e.g., swapping only the $x$-coordinates but not $y$), leading to an incorrect subtrahend.
\item \textbf{-1}: Stating the formula $\binom{8}{4}^2 - \binom{10}{4}\binom{6}{2}$ correctly but failing to provide \textbf{any} justification (neither naming the theorem/lemma nor describing the bijection). Deduct from validity/strategy points.
\end{itemize}
\end{enumerate}
\endgroup
\end{evalbox}

\section{Graded Examples} \label{app:graded_examples}

This section contains all problem prompts, human evaluations, and model solutions from the blog post.

\subsection{PB-Basic-001}
\begin{promptbox}{Problem}
\begingroup
\small
\setlength{\parindent}{0pt}
Determine all functions $f: \mathbb{Z} \rightarrow \mathbb{Z}$ such that, for all $x, y \in \mathbb{Z}$, we have \[ f(2x)+2f(y)=f(f(x+y)).\]
\endgroup
\end{promptbox}
\begin{evalbox}{Human Evaluation: QED-Nano (7/7)}
\begingroup
\small
\setlength{\parindent}{0pt}
The proof is correct and very readable. It explicitly derives the key identities early (e.g., substituting special values to obtain $f(2x) = 2f(x) - c$). \newline
It fully expands its algebraic manipulations, maybe a bit too extensively, but this makes verification easy despite the extra length.
\endgroup
\end{evalbox}
\begin{solutionbox}{Model Solution: QED-Nano (7/7)}
\begingroup
\small
\setlength{\parindent}{0pt}
\textbf{Solution}

Let \(f:\mathbb Z\to\mathbb Z\) satisfy

\[ \newline
f(2x)+2f(y)=f\bigl(f(x+y)\bigr)\qquad(\forall x,y\in\mathbb Z). \tag{1} \newline
\]

\#\#\# 1.  First consequences

Put \(x=0\) in (1):

\[ \newline
f(0)+2f(y)=f\bigl(f(y)\bigr)\;\Longrightarrow\; f(f(y))=f(0)+2f(y). \tag{2} \newline
\]

Put \(y=0\) in (1):

\[ \newline
f(2x)+2f(0)=f\bigl(f(x)\bigr). \newline
\]

Using (2) with argument \(x\) gives \(f(f(x))=f(0)+2f(x)\); hence

\[ \newline
f(2x)+2f(0)=f(0)+2f(x)\;\Longrightarrow\; f(2x)=2f(x)-f(0). \tag{3} \newline
\]

Denote \(c:=f(0)\); then (3) becomes

\[ \newline
f(2x)=2f(x)-c\qquad(\forall x\in\mathbb Z). \tag{3'} \newline
\]

\#\#\# 2.  A Cauchy-type relation

Insert (3') into (1):

\[ \newline
2f(x)-c+2f(y)=f\bigl(f(x+y)\bigr). \tag{4} \newline
\]

By (2) with \(y\) replaced by \(x+y\) we have \(f(f(x+y))=c+2f(x+y)\). \newline
Thus (4) yields

\[ \newline
2f(x)+2f(y)-c=c+2f(x+y)\;\Longrightarrow\;2f(x)+2f(y)-2c=2f(x+y). \newline
\]

Because \(2A=0\) in \(\mathbb Z\) forces \(A=0\), we can divide by \(2\) and obtain

\[ \newline
f(x+y)=f(x)+f(y)-c\qquad(\forall x,y\in\mathbb Z). \tag{5} \newline
\]

\#\#\# 3.  An additive function

Define \(g(x):=f(x)-c\). Then \(f(x)=g(x)+c\) and (5) becomes

\[ \newline
g(x+y)+c=(g(x)+c)+(g(y)+c)-c=g(x)+g(y)+c, \newline
\]

so

\[ \newline
g(x+y)=g(x)+g(y)\qquad(\forall x,y\in\mathbb Z). \tag{6} \newline
\]

Hence \(g\) is additive on \(\mathbb Z\).

\#\#\# 4.  Structure of additive functions on \(\mathbb Z\)

For an additive function \(g:\mathbb Z\to\mathbb Z\) we have \(g(0)=0\) (from \(g(0)=g(0+0)=g(0)+g(0)\)). \newline
By induction \(g(n)=n\,g(1)\) for all positive integers \(n\); moreover \(g(-x)=-g(x)\) because \(g(x)+g(-x)=g(0)=0\). \newline
Consequently \(g(k)=k\,g(1)\) for every integer \(k\).

Set \(k:=g(1)\) (an integer). Then for all \(x\in\mathbb Z\),

\[ \newline
g(x)=k\,x. \tag{7} \newline
\]

\#\#\# 5.  Returning to \(f\) and using the remaining condition

From (7) we obtain \(f(x)=g(x)+c=kx+c\).

Now insert this form into (2). Compute

\[ \newline
f(f(y))=f(ky+c)=k(ky+c)+c=k^{2}y+kc+c, \newline
\]

while (2) gives

\[ \newline
f(f(y))=c+2f(y)=c+2(ky+c)=2ky+3c. \newline
\]

Equating the two expressions yields

\[ \newline
k^{2}y+kc+c=2ky+3c\qquad(\forall y\in\mathbb Z). \newline
\]

Rearranging,

\[ \newline
(k^{2}-2k)y+(kc-2c)=0\qquad(\forall y\in\mathbb Z). \tag{8} \newline
\]

\#\#\# 6.  Solving for \(k\) and \(c\)

Equation (8) is an identity in the variable \(y\). Being an integer-valued linear polynomial that vanishes for all integers, its coefficients must be zero:

\[ \newline
k^{2}-2k=0,\qquad kc-2c=0. \newline
\]

The first gives \(k(k-2)=0\); hence \(k=0\) or \(k=2\).

* If \(k=0\), then \(kc-2c=-2c=0\) forces \(c=0\). Thus \(f(x)=0\) for all \(x\). \newline
* If \(k=2\), the second condition is automatically satisfied (\(2c-2c=0\)), and \(c\) may be any integer. Hence \(f(x)=2x+c\).

\#\#\# 7.  Verification

Both families indeed satisfy the original equation.

* For \(f\equiv0\) both sides are clearly \(0\). \newline
* For \(f(x)=2x+c\),

\[ \newline
f(2x)+2f(y)=(4x+c)+2(2y+c)=4x+4y+3c, \newline
\] \newline
\[ \newline
f\bigl(f(x+y)\bigr)=f\bigl(2(x+y)+c\bigr)=2\bigl(2(x+y)+c\bigr)+c=4x+4y+3c. \newline
\]

Thus equality holds.

\#\#\# 8.  Conclusion

All functions \(f:\mathbb Z\to\mathbb Z\) satisfying

\[ \newline
f(2x)+2f(y)=f\bigl(f(x+y)\bigr)\qquad(\forall x,y\in\mathbb Z) \newline
\]

are exactly

\[ \newline
\boxed{\;f(x)\equiv0\;}\qquad\text{or}\qquad\boxed{\;f(x)=2x+d\;}\ (\text{with an arbitrary integer }d). \newline
\]
\endgroup
\end{solutionbox}
\begin{evalbox}{Human Evaluation: QED-Nano (Agent) (7/7)}
\begingroup
\small
\setlength{\parindent}{0pt}
This follows nearly the same structure as the proof of the base model. However, one notable artifact is the extra self-evaluation block at the end, which is due to the prompt used in the DS-Math agent.
\endgroup
\end{evalbox}
\begin{solutionbox}{Model Solution: QED-Nano (Agent) (7/7)}
\begingroup
\small
\setlength{\parindent}{0pt}
\textbf{Solution}

Let \(f:\mathbb Z\to\mathbb Z\) satisfy

\[ \newline
f(2x)+2f(y)=f\bigl(f(x+y)\bigr)\qquad(\forall x,y\in\mathbb Z). \tag{0} \newline
\]

\#\#\# 1.  Simple consequences of (0)

Put \(x=0\) in (0):

\[ \newline
f(0)+2f(y)=f(f(y))\tag{1} \newline
\]

Put \(y=0\) in (0):

\[ \newline
f(2x)+2f(0)=f(f(x)).\tag{2} \newline
\]

From (1) we obtain for every integer \(t\)

\[ \newline
f(f(t))=f(0)+2f(t).\tag{3} \newline
\]

Equating (2) and (3) gives

\[ \newline
f(2x)+2f(0)=f(0)+2f(x)\;\Longrightarrow\;f(2x)=2f(x)-f(0).\tag{4} \newline
\]

\#\#\# 2.  A Cauchy-type equation

Insert (4) and (3) into (0):

\[ \newline
f(2x)+2f(y)=f(0)+2f(x+y) \newline
\;\Longrightarrow\;(2f(x)-f(0))+2f(y)=f(0)+2f(x+y). \newline
\]

After simplification

\[ \newline
2f(x)+2f(y)-2f(0)=2f(x+y). \newline
\]

Since \(2\) is not a zero-divisor in \(\mathbb Z\), we may cancel it and obtain

\[ \newline
f(x)+f(y)-f(0)=f(x+y)\qquad(\forall x,y\in\mathbb Z).\tag{5} \newline
\]

\#\#\# 3.  Reducing to an additive function

Set \(g(x)=f(x)-f(0)\); then \(g(0)=0\) and (5) becomes

\[ \newline
(g(x)+f(0))+(g(y)+f(0))-f(0)=g(x+y)+f(0) \newline
\;\Longrightarrow\;g(x)+g(y)=g(x+y). \newline
\]

Thus \(g\) is additive: \(g(x+y)=g(x)+g(y)\) for all integers \(x,y\).

For additive functions on \(\mathbb Z\) one has \(g(n)=n\,g(1)\) for \(n\ge0\) (induction) and \(g(-n)=-g(n)\); hence

\[ \newline
g(x)=c\,x\quad\text{with }c=g(1)\in\mathbb Z. \newline
\]

Consequently

\[ \newline
f(x)=g(x)+f(0)=c\,x+b,\qquad b:=f(0)\in\mathbb Z.\tag{6} \newline
\]

\#\#\# 4.  Determining the admissible constants

Substitute \(f(x)=c x+b\) into the original equation (0).

\[ \newline
\begin{aligned} \newline
\text{LHS}&=f(2x)+2f(y)=c(2x)+b+2(c y+b)=2c(x+y)+3b,\\[2mm] \newline
\text{RHS}&=f\bigl(f(x+y)\bigr)=f\bigl(c(x+y)+b\bigr)=c\bigl(c(x+y)+b\bigr)+b \newline
            =c^{2}(x+y)+c b+b. \newline
\end{aligned} \newline
\]

Equality gives

\[ \newline
2c(x+y)+3b=c^{2}(x+y)+c b+b. \newline
\]

Rearranging:

\[ \newline
\bigl(2c-c^{2}\bigr)(x+y)+\bigl(3b-c b-b\bigr)=0 \newline
\;\Longrightarrow\;c(2-c)(x+y)+b(2-c)=0. \newline
\]

Factorising \((2-c)\) yields

\[ \newline
(2-c)\bigl(c(x+y)+b\bigr)=0\qquad(\forall x,y\in\mathbb Z). \tag{7} \newline
\]

Because \(x+y\) can be any integer (choose \(y=t-x\)), (7) is equivalent to

\[ \newline
(2-c)(c t+b)=0\quad\text{for all }t\in\mathbb Z. \tag{8} \newline
\]

\#\#\# 5.  Solving (8)

* If \(c=2\), then \(2-c=0\) and (8) holds for any integer \(b\). \newline
  Hence \(f(x)=2x+b\) is a solution.

* If \(c\neq2\), then \(2-c\neq0\); therefore we must have \(c t+b=0\) for every integer \(t\). \newline
  A linear function \(h(t)=c t+b\) that vanishes on all integers must satisfy \(c=0\) (otherwise \(h(t)\) takes infinitely many values) and then \(b=h(0)=0\). \newline
  Thus \(c=0,\;b=0\), i.e. \(f(x)\equiv0\).

\#\#\# 6.  Conclusion

All functions \(f:\mathbb Z\to\mathbb Z\) satisfying

\[ \newline
f(2x)+2f(y)=f\bigl(f(x+y)\bigr)\quad(\forall x,y\in\mathbb Z) \newline
\]

are exactly

\[ \newline
\boxed{\,f(x)=0\ \text{for all }x,\quad\text{or}\quad f(x)=2x+b\ \text{with a fixed integer }b\,}. \newline
\]

A direct verification confirms that both families indeed satisfy the given condition.

Evaluation

The solution is completely correct. Every step is justified, the case analysis is thorough, and the final characterization covers all possibilities without omissions. No hidden assumptions or mistakes are present. Therefore the score is 1.
\endgroup
\end{solutionbox}
\begin{evalbox}{Human Evaluation: Qwen3-4B-Think (1/7)}
\begingroup
\small
\setlength{\parindent}{0pt}
The model does not actually prove the statement. It derives a few preliminary identities and then immediately attempts to guess the answer by trying linear functions. While it gives the correct answer, it is completely invalid as a proof.
\endgroup
\end{evalbox}
\begin{solutionbox}{Model Solution: Qwen3-4B-Think (1/7)}
\begingroup
\small
\setlength{\parindent}{0pt}
We are given a functional equation over the integers:

$$ \newline
f(2x) + 2f(y) = f(f(x + y)) \quad \text{for all } x, y \in \mathbb{Z} \newline
$$

Our goal is to determine all functions $ f : \mathbb{Z} \to \mathbb{Z} $ satisfying this condition.

\#\#\# Step 1: Initial Substitutions and Key Observations

Let’s define $ c = f(0) $. We begin by setting $ x = 0 $ in the original equation:

$$ \newline
f(0) + 2f(y) = f(f(y)) \Rightarrow f(f(y)) = 2f(y) + c \quad \text{(Equation A)} \newline
$$

Next, set $ y = 0 $:

$$ \newline
f(2x) + 2f(0) = f(f(x)) \Rightarrow f(f(x)) = f(2x) + 2c \quad \text{(Equation B)} \newline
$$

Equating (A) and (B), we get:

$$ \newline
f(2x) + 2c = 2f(x) + c \Rightarrow f(2x) = 2f(x) - c \quad \text{(Equation C)} \newline
$$

This gives us a recursive-like relation for $ f $ at even arguments.

\#\#\# Step 2: Try Linear Functions

Assume $ f(x) = ax + b $, with $ a, b \in \mathbb{Z} $. Plug into the original equation:

- Left-hand side: $ f(2x) + 2f(y) = 2a x + b + 2(a y + b) = 4x + 4y + 3b $ \newline
- Right-hand side: $ f(f(x+y)) = f(a(x+y) + b) = a(a(x+y) + b) + b = a^2(x+y) + ab + b $

Set both sides equal:

$$ \newline
4x + 4y + 3b = a^2(x + y) + ab + b \newline
$$

Matching coefficients: \newline
- Coefficient of $ x $: $ 4 = a^2 \Rightarrow a = \pm 2 $ \newline
- Constant term: $ 3b = ab + b \Rightarrow 2b = ab \Rightarrow b(a - 2) = 0 $

So: \newline
- If $ a = 2 $, then $ b $ can be any integer. \newline
- If $ a = 0 $, then $ b = 0 $.

Thus, the only linear solutions are: \newline
- $ f(x) = 2x + c $ for some $ c \in \mathbb{Z} $ \newline
- $ f(x) = 0 $ (the zero function)

We will now prove that these are the only solutions.

\#\#\# Step 3: General Analysis Using Functional Properties

Define $ h(x) = f(x) - 2x $. Then $ f(x) = h(x) + 2x $, and substituting into the original equation leads to the following key properties of $ h $:

1. $ h $ is periodic with period $ c = f(0) = h(0) $. \newline
2. For all $ z \in \text{im}(f) $, $ h(z) = c $. \newline
3. $ h(2x) = 2h(x) - c $ for all $ x \in \mathbb{Z} $

From these, we derive:

- If $ c \neq 0 $, then $ h(x) = c $ for all $ x $, so $ f(x) = 2x + c $ \newline
- If $ c = 0 $, then $ h(x) $ satisfies $ h(2x) = 2h(x) $ and $ h(f(x)) = 0 $. This forces $ h(x) = 0 $ or $ h(x) = -2x $, leading to: \newline
  - $ f(x) = 2x $ (from $ h(x) = 0 $) \newline
  - $ f(x) = 0 $ (from $ h(x) = -2x $)

Hence, the only solutions are:

- The zero function: $ f(x) = 0 $ for all $ x \in \mathbb{Z} $ \newline
- All functions of the form $ f(x) = 2x + c $ for some fixed integer $ c $

These are distinct because $ f(x) = 0 $ cannot be written as $ 2x + c $ for any $ c \in \mathbb{Z} $.

\#\#\# Final Answer

$$ \newline
\boxed{ \newline
f(x) = 0 \quad \text{for all } x \in \mathbb{Z}, \quad \text{or} \quad f(x) = 2x + c \quad \text{for some } c \in \mathbb{Z} \newline
} \newline
$$
\endgroup
\end{solutionbox}
\begin{evalbox}{Human Evaluation: Nomos-1 (7/7)}
\begingroup
\small
\setlength{\parindent}{0pt}
The proof is correct and very compact. There is a minor typo when it mentions that "$y$ should be replaced by $y$", but the equation that follows is correct and can be easily derived. Relative to the other models it is less computational and more concise, which is fine here because the logical dependencies are still explicit enough to verify.
\endgroup
\end{evalbox}
\begin{solutionbox}{Model Solution: Nomos-1 (7/7)}
\begingroup
\small
\setlength{\parindent}{0pt}
The functions satisfying \(f(2x)+2f(y)=f(f(x+y))\) for all \(x,y\in\mathbb{Z}\) are exactly the zero function and all functions of the form \(f(x)=2x+k\) with \(k\in\mathbb{Z}\).

Proof. \newline
Let \(P(x,y)\) denote the given equation.

Set \(x=0\) in \(P(x,y)\): \newline
\[ \newline
f(0)+2f(y)=f(f(y))\qquad\text{for all }y\in\mathbb{Z}. \tag{1} \newline
\]

Set \(y=0\) in \(P(x,y)\): \newline
\[ \newline
f(2x)+2f(0)=f(f(x))\qquad\text{for all }x\in\mathbb{Z}. \tag{2} \newline
\]

Replacing \(x\) by \(y\) in (2) gives \(f(f(y))=f(2y)+2f(0)\).  Equating this with (1) yields \newline
\[ \newline
f(0)+2f(y)=f(2y)+2f(0)\;\Longrightarrow\; f(2y)=2f(y)-f(0)\quad\text{for all }y. \tag{3} \newline
\]

Now apply (1) with \(y=x+y\): \newline
\[ \newline
f(f(x+y))=f(0)+2f(x+y). \tag{4} \newline
\]

Substitute (4) and (3) (with \(y\) replaced by \(y\)) into the original equation: \newline
\[ \newline
f(0)+2f(x+y)=f(2x)+2f(y)=\bigl(2f(x)-f(0)\bigr)+2f(y). \newline
\] \newline
Hence \newline
\[ \newline
2f(x+y)=2f(x)+2f(y)-2f(0)\;\Longrightarrow\; f(x+y)=f(x)+f(y)-f(0)\quad\text{for all }x,y. \tag{5} \newline
\]

Define \(g(x)=f(x)-f(0)\). Then \(g(0)=0\) and (5) becomes \newline
\[ \newline
g(x+y)+f(0)=\bigl(g(x)+f(0)\bigr)+\bigl(g(y)+f(0)\bigr)-f(0)\;\Longrightarrow\; g(x+y)=g(x)+g(y). \newline
\] \newline
Thus \(g:\mathbb{Z}\to\mathbb{Z}\) is additive.  For integers, any additive function is linear: \(g(x)=cx\) with \(c=g(1)\in\mathbb{Z}\).  Consequently, \newline
\[ \newline
f(x)=cx+d,\qquad\text{where }d=f(0). \tag{6} \newline
\]

Insert (6) into the original equation: \newline
\[ \newline
f(2x)+2f(y)=c(2x)+d+2(cy+d)=2cx+2cy+3d, \newline
\] \newline
\[ \newline
f(f(x+y))=c\bigl(c(x+y)+d\bigr)+d=c^2(x+y)+cd+d. \newline
\] \newline
Equality for all \(x,y\) forces \newline
\[ \newline
2c=c^2\quad\text{and}\quad 3d=cd+d. \newline
\]

- \(2c=c^2\) gives \(c(c-2)=0\), so \(c=0\) or \(c=2\). \newline
- If \(c=0\) then \(3d=d\) implies \(2d=0\), i.e. \(d=0\). \newline
- If \(c=2\) the constant condition \(3d=2d+d\) holds for any \(d\).

Thus the only possibilities are

* \(c=0,\;d=0\) giving \(f(x)=0\) for all \(x\); \newline
* \(c=2\) with arbitrary integer \(d\), giving \(f(x)=2x+d\).

A direct check shows that both families satisfy the original functional equation. Hence these are all solutions.
\endgroup
\end{solutionbox}

\subsection{PB-Basic-004}
\begin{promptbox}{Problem}
\begingroup
\small
\setlength{\parindent}{0pt}
Find all strictly increasing functions $g:\mathbb{R} \to \mathbb{R}$ such that: \newline
 (a) $g$ is surjective \newline
 (b) $g(g(x))=g(x)+20x.$
\endgroup
\end{promptbox}
\begin{evalbox}{Human Evaluation: QED-Nano (1/7)}
\begingroup
\small
\setlength{\parindent}{0pt}
The proof does make nontrivial early progress (it establishes some useful preliminary properties of $g$ and converts it to a problem in $f(x) = g(x) - 5x$), but it then fails at a central transition: in the recurrence step it makes a miscalculation by saying that $x_{n+1}-x_n = f(x_n)$, which is not true. That mismatch is critical, because the rest of the argument depends on that identity being exact.
\endgroup
\end{evalbox}
\begin{evalbox}{Human Evaluation: QED-Nano (Agent) (6/7)}
\begingroup
\small
\setlength{\parindent}{0pt}
Compared with the base model, this proof avoids the earlier fatal recurrence error and is now (mostly) correct. It is the only time we have seen the model produce a separate lemma, and funnily enough it introduces a mistake: in the statement of the lemma it solely requires that $\psi$ maps to $\mathbb{R}$, but several steps in the proof require that it maps to $(1,\infty)$. It is possible to make this assumption since $\Phi(x) > 1$. The LLM judge either missed it or considered it acceptable, but one point should be deducted for the oversight.
\endgroup
\end{evalbox}
\begin{evalbox}{Human Evaluation: Qwen3-4B-Think (1/7)}
\begingroup
\small
\setlength{\parindent}{0pt}
The writeup contains irrelevant argument and a circular core argument ("Assume X. We are not able to derive that it does not hold. Therefore, $X$ holds.).
\endgroup
\end{evalbox}
\begin{evalbox}{Human Evaluation: Nomos-1 (7/7)}
\begingroup
\small
\setlength{\parindent}{0pt}
This proof is quite compact, but uses a similar technique to the DS agent (albeit a tad bit more complex). It is harder to read because many intermediate motivations are omitted, making it take quite some time to verify. Statements like "A direct computation gives", would be acceptable if it came from an expert mathematician, but from LLMs it makes things much more difficult to actually verify.
\endgroup
\end{evalbox}

\subsection{PB-Basic-005}
\begin{promptbox}{Problem}
\begingroup
\small
\setlength{\parindent}{0pt}
Let $P$ be a polynomial with real coefficients whose leading coefficient is $1$. Suppose that for all nonzero real numbers $x$, we have $P(x) + P(1/x) = \frac{P(x + 1/x) + P(x - 1/x)}{2}$. Determine all possibilities for $P$.
\endgroup
\end{promptbox}
\begin{evalbox}{Human Evaluation: QED-Nano (4/7)}
\begingroup
\small
\setlength{\parindent}{0pt}
The model first correctly shows that $P$ is even and follows the same broad strategy as the official solution. The problem appears in the coefficient-extraction phase: when isolating the $x^{2m-4}$ term, the argument assumes this contribution is absent in $\sum_{k=0}^{m} a_k x^{-2k}$ for $m=2$, which is only true for $m > 2$. Thereby, it misses valid cases. So the method is mostly on the right track, but it misses solutions because of that indexing/coefficient oversight. Given that the final set is incomplete, the original high score by the LLM judge (6/7) is too generous. Importantly, compared to the official solution it is also less shy of full computations: rather than solely trying to obtain the coefficient with $x^{2m-4}$, it first computes the full polynomial, which is unnecessary.
\endgroup
\end{evalbox}
\begin{evalbox}{Human Evaluation: QED-Nano (Agent) (7/7)}
\begingroup
\small
\setlength{\parindent}{0pt}
The DS-agent version fixes the base model's missing-case issue and completes the coefficient analysis correctly. While the necessary fix was rather minimal, it does introduce a lot of additional computations to get around it fully. It makes the proof quite tedious and a bit difficult to follow, but it is correct.
\endgroup
\end{evalbox}
\begin{evalbox}{Human Evaluation: Qwen3-4B-Think (1/7)}
\begingroup
\small
\setlength{\parindent}{0pt}
The model correctly identifies that $P$ is even, but then relies on testing small examples and informally extrapolating to the full answer. There is no complete coefficient-level proof covering all cases, so this is not a valid solution. While this proof is much worse than QED-Nano, it does find the correct answer in contrast to QED-Nano.
\endgroup
\end{evalbox}
\begin{evalbox}{Human Evaluation: Nomos-1 (0/7)}
\begingroup
\small
\setlength{\parindent}{0pt}
The model loops in its CoT and uses up its 260,000 token window.
\endgroup
\end{evalbox}

\subsection{PB-Basic-024}
\begin{promptbox}{Problem}
\begingroup
\small
\setlength{\parindent}{0pt}
Find all non-negative integers $a,b,c$ such that $20^{a}+b^{4}=2024^{c}$
\endgroup
\end{promptbox}
\begin{evalbox}{Human Evaluation: QED-Nano (0/7)}
\begingroup
\small
\setlength{\parindent}{0pt}
No usable proof is produced. The chain of thought enters a repetition loop (repeating the same equality) and never reaches a mathematical argument for the claim.
\endgroup
\end{evalbox}
\begin{evalbox}{Human Evaluation: QED-Nano (Agent) (7/7)}
\begingroup
\small
\setlength{\parindent}{0pt}
Unlike the base model (which get stuck in an infinite loop), the DS agent gives a correct proof. In contrast to the ground-truth human solution, the model uses a more computational approach that is quite direct and simple, but for humans would (almost) require a computer to ensure no computational errors are made.
\endgroup
\end{evalbox}
\begin{evalbox}{Human Evaluation: Qwen3-4B-Think (7/7)}
\begingroup
\small
\setlength{\parindent}{0pt}
This is a complete proof and closely mirrors the successful DS-agent method.
\endgroup
\end{evalbox}
\begin{evalbox}{Human Evaluation: Nomos-1 (7/7)}
\begingroup
\small
\setlength{\parindent}{0pt}
Clean and short proof. It uses the same modular-structure insight as the other correct solutions but in a tighter presentation. The brevity does not hide key steps in this case, the logic is still checkable.
\endgroup
\end{evalbox}

\subsection{PB-Basic-028}
\begin{promptbox}{Problem}
\begingroup
\small
\setlength{\parindent}{0pt}
In $\triangle ABC$ the altitudes $BE$ and $CF$ intersect at $H$. A circle $(W)$ is \newline
 externally tangent to the Euler circle $(E)$ of $\triangle ABC$ and also tangent \newline
 to the sides $AB$ and $AC$ at $X$ and $Y$, respectively, with \newline
 $(W)$ being closer to $A$ than the Euler circle. Let $I'$ be the \newline
 incenter of $\triangle AEF$. Prove that $AXI'Y$ is a rhombus.
\endgroup
\end{promptbox}
\begin{evalbox}{Human Evaluation: QED-Nano (0/7)}
\begingroup
\small
\setlength{\parindent}{0pt}
This is a coordinate-bash attempt, which is acceptable in principle, but in the end it requires the assumption that $\angle BAC=60^\circ$ to be correct, which is not true in general. The LLM judge identifies a specific computational mistake, and the endpoint clearly contradicts the intended universal claim, so this should be graded as incorrect, and not as 1/7 as done by the LLM judge.
\endgroup
\end{evalbox}
\begin{evalbox}{Human Evaluation: QED-Nano (Agent) (7/7)}
\begingroup
\small
\setlength{\parindent}{0pt}
This remains a coordinate-heavy solution, but unlike the failed base model attempt it appears to repair the specific algebraic mistake and now reaches the correct claim rather than a restricted-angle condition. The derivation is long and not elegant, and given its complexity I will simply trust the LLM judge that it is correct.
\endgroup
\end{evalbox}
\begin{evalbox}{Human Evaluation: Qwen3-4B-Think (0/7)}
\begingroup
\small
\setlength{\parindent}{0pt}
The proof uses geometric assumptions (i.e., it assumes the triangle is isosceles) that are not implied by the problem. Those assumptions may work in some answer-finding settings, but they invalidate any proof.
\endgroup
\end{evalbox}
\begin{evalbox}{Human Evaluation: Nomos-1 (0/7)}
\begingroup
\small
\setlength{\parindent}{0pt}
The model loops in its CoT and uses up its 260,000 token window.
\endgroup
\end{evalbox}

\subsection{PB-Advanced-013}
\begin{promptbox}{Problem}
\begingroup
\small
\setlength{\parindent}{0pt}
For an integer $n \geq 2$, let $a_{1} \leq a_{2} \leq \cdots \leq a_{n}$ be positive real numbers satisfying $a_{1} a_{2} \cdots a_{n}=1$. For each $k=1,2, \cdots, n$, define $b_{k}=2^{k}\left(1+a_{k}^{2^{k}}\right)$. Prove that the following inequality holds:

 \[ \newline
 \frac{1}{2}-\frac{1}{2^{n+1}} \leq \frac{1}{b_{1}}+\frac{1}{b_{2}}+\cdots+\frac{1}{b_{n}} \newline
 \]
\endgroup
\end{promptbox}
\begin{evalbox}{Human Evaluation: QED-Nano (6/7)}
\begingroup
\small
\setlength{\parindent}{0pt}
The approach is very different from the human solution: it uses a Lagrange-multiplier optimization route rather than the inequality+induction route. While correct, Langrange multipliers are typically not high-school material and would therefore not be used by human participants. It is, however, more intuitive and straightforward: the Lagrange method essentially guarantees that you will be able to prove the statement, while the human approach (which is less computational) relies on a more arbitrary inequality that is not directly clear. There is a small problem when noting that $t_k$ is non-decreasing (it does not follow from the statement made, but it does hold in this case).
\endgroup
\end{evalbox}
\begin{evalbox}{Human Evaluation: QED-Nano (Agent) (0/7)}
\begingroup
\small
\setlength{\parindent}{0pt}
This is a actually worse than the base model. The proof applies Jensen to $\operatorname{arctanh}$. This requires the function to be convex on $(-1,1)$, which is false. Interestingly, $\operatorname{arctanh}$ is a very well-known function, so the verifier should have noticed and pointed out the issue, but it apparently missed it. To show it is convex, the model computes the first derivative instead of the second (claiming it is the second derivative), which is quite sneaky.
\endgroup
\end{evalbox}

\subsection{PB-Advanced-025}
\begin{promptbox}{Problem}
\begingroup
\small
\setlength{\parindent}{0pt}
Let $k$ and $d$ be positive integers. Prove that there exists a \newline
 positive integer $N$ such that for every odd integer $n>N$, the \newline
 digits in the base-$2n$ representation of $n^{k}$ are all greater \newline
 than $d$.
\endgroup
\end{promptbox}
\begin{evalbox}{Human Evaluation: QED-Nano (7/7)}
\begingroup
\small
\setlength{\parindent}{0pt}
The proof is correct and easy to verify. Compared to the official solution it is more computation-heavy, but each transformation is explicit making verification easier.
\endgroup
\end{evalbox}
\begin{evalbox}{Human Evaluation: QED-Nano (Agent) (7/7)}
\begingroup
\small
\setlength{\parindent}{0pt}
The proof is essentially the same successful strategy as the base model, with only small differences.
\endgroup
\end{evalbox}

\subsection{PB-Advanced-030}
\begin{promptbox}{Problem}
\begingroup
\small
\setlength{\parindent}{0pt}
Let $m$ and $n$ be positive integers with $m\ge n$. There are \newline
 $m$ cupcakes of different flavors arranged around a circle and $n$ \newline
 people who like cupcakes. Each person assigns a non-negative real \newline
 number score to each cupcake, depending on how much they like the \newline
 cupcake. Suppose that for each person $P$, it is possible to partition \newline
 the circle of $m$ cupcakes into $n$ groups of consecutive cupcakes \newline
 so that the sum of $P$'s scores of the cupcakes in each group is \newline
 at least $1$. Prove that it is possible to distribute the $m$ cupcakes \newline
 to the $n$ people so that each person $P$ receives cupcakes of total \newline
 score at least $1$ with respect to $P$.
\endgroup
\end{promptbox}
\begin{evalbox}{Human Evaluation: QED-Nano (0/7)}
\begingroup
\small
\setlength{\parindent}{0pt}
The proof relies on more advanced mathematics (measure-theoretic framing, Borel-language, necklace-splitting style arguments), that go beyond the high-school level and the necklace splitting theorem is even somewhat niche for bachelor students. Its use is flawed in this scenario though, since the probability measures are not continuous, and, more importantly, because it assumes that $s_i(p)$ is the same for all $i$ in the final argument.
\endgroup
\end{evalbox}
\begin{evalbox}{Human Evaluation: QED-Nano (Agent) (0/7)}
\begingroup
\small
\setlength{\parindent}{0pt}
This repeats the same invalid high-level route as the base model and does not fix the essential issues.
\endgroup
\end{evalbox}

\subsection{IMO-2025-Q1}
\begin{promptbox}{Problem}
\begingroup
\small
\setlength{\parindent}{0pt}
A line in the plane is called sunny \newline
if it is not parallel to any of the $x$-axis, the $y$-axis, or the line $x+y=0$.

Let $n \ge 3$ be a given integer. \newline
Determine all nonnegative integers $k$ such that there exist $n$ distinct lines \newline
in the plane satisfying both of the following:

(ii) for all positive integers $a$ and $b$ with $a+b\le n+1$, \newline
  the point $(a,b)$ lies on at least one of the lines; and \newline
(ii) exactly $k$ of the $n$ lines are sunny.
\endgroup
\end{promptbox}
\begin{evalbox}{Human Evaluation: QED-Nano (0/7)}
\begingroup
\small
\setlength{\parindent}{0pt}
The answer is incorrect, and the argument can therefore not be correct either. \newline
The LLM-judge correctly follows the schema in the ProofBench data, but this schema is too lenient. \newline
No points should be given for this solution according to the schema from the MathArena benchmark.
\endgroup
\end{evalbox}
\begin{evalbox}{Human Evaluation: QED-Nano (Agent) (0/7)}
\begingroup
\small
\setlength{\parindent}{0pt}
The answer is incorrect, and the argument can therefore not be correct either.
\endgroup
\end{evalbox}

\subsection{IMO-2025-Q2}
\begin{promptbox}{Problem}
\begingroup
\small
\setlength{\parindent}{0pt}
Let $\Omega$ and $\Gamma$ be circles with centres $M$ and $N$, respectively, \newline
such that the radius of $\Omega$ is less than the radius of $\Gamma$. \newline
Suppose $\Omega$ and $\Gamma$ intersect at two distinct points $A$ and $B$. \newline
Line $MN$ intersects $\Omega$ at $C$ and $\Gamma$ at $D$, \newline
so that $C$, $M$, $N$, $D$ lie on $MN$ in that order. \newline
Let $P$ be the circumcenter of triangle $ACD$. \newline
Line $AP$ meets $\Omega$ again at $E \neq A$ and meets $\Gamma$ again at $F \neq A$. \newline
Let $H$ be the orthocenter of triangle $PMN$.

Prove that the line through $H$ parallel to $AP$ is tangent \newline
to the circumcircle of triangle $BEF$.
\endgroup
\end{promptbox}
\begin{evalbox}{Human Evaluation: QED-Nano (6/7)}
\begingroup
\small
\setlength{\parindent}{0pt}
There are several minor mistakes: \newline
- Equation 6.2 is wrong as stated, but is likely a typo by the model, as it uses the correct form (without $-y$) in the remainder of its proof. \newline
- Equation 7.1 should be multiplied by L on the RHS. \newline
- Between 7.1 and 7.2, they divide by an extra factor $L$. \newline
- Equation 7.2 should divide $p\alpha$ by $L$ as well in the LHS. \newline
- Equation 11.1 should contain $y^2$ in the denominator, but it uses the correct form in the remainder of its proof. \newline
In this case, the problems are quite minor and could be very easily fixed. \newline
Therefore, I give a 6/7.
\endgroup
\end{evalbox}
\begin{evalbox}{Human Evaluation: QED-Nano (Agent) (0/7)}
\begingroup
\small
\setlength{\parindent}{0pt}
The proof is very hard to verify since it is a fully bashed solution. \newline
Further, despite the proof already being lengthy in computations, it frequently skips major computational steps (e.g., at some point saying "A straightforward (though lengthy) algebraic simplification shows that"). \newline
Unfortunately, despite the LLM judge claiming the solution is correct, it is not. \newline
There are two issues: \newline
(1) Initially, D is a point. The model then redefines it to be the constant $\gamma^2 + \Delta^2$. It then later uses that $D^2$ equals $\gamma^2 + \Delta^2$, which is obviously wrong. \newline
(2) Expression (18) is not correct, it should be $T(S + y_0^{2}T)/(4\Delta^2) - \frac{S\gamma}{(2y_0)} - T y_0 \gamma + \Delta^2 \gamma^2$ \newline
Given that their are (1) algebraic errors in the proof, (2) the proof does not give interesting insights, and (3) the proof skips major computational steps that are non-trivial to complete, the proof should be graded as 0, not 7.
\endgroup
\end{evalbox}

\subsection{IMO-2025-Q3}
\begin{promptbox}{Problem}
\begingroup
\small
\setlength{\parindent}{0pt}
A function $f \colon \mathbb{N} \to \mathbb{N}$ is said to be bonza if \newline
\[ f(a)\quad\text{divides}\quad b^a-f(b)^{f(a)} \] \newline
for all positive integers $a$ and $b$.

Determine the smallest real constant $c$ such that $f(n) \leq cn$ \newline
for all bonza functions $f$ and all positive integers $n$.
\endgroup
\end{promptbox}
\begin{evalbox}{Human Evaluation: QED-Nano (1/7)}
\begingroup
\small
\setlength{\parindent}{0pt}
The proof often asserts statements, rather than proof them. \newline
For instance, the model asserts their example is Bonza, which, while true, needs to be verified rather than stated. \newline
The use of Zsigmondy's theorem is false, as correctly pointed out by the LLM judge. \newline
The case of uneven $n$ when $f(2) \in \{1,2\}$ is also just asserted without providing any justification. \newline
The proof gets one point for providing the correct example showing that $c=4$.
\endgroup
\end{evalbox}
\begin{evalbox}{Human Evaluation: QED-Nano (Agent) (2/7)}
\begingroup
\small
\setlength{\parindent}{0pt}
The construction for the bound $c=4$ is correct, everything else is incorrect. \newline
In particular, it ignores that everything divides $0$ when proving that $f(3)=1$, which is essential for the remainder of the proof. \newline
It also hallucinates several results, including "a theorem of Bang-Egner-Zsigmondy" (which I could not find). \newline
The proof itself is a bit all over the place, clearly indicating the model was struggling with the problem.
\endgroup
\end{evalbox}

\subsection{IMO-2025-Q4}
\begin{promptbox}{Problem}
\begingroup
\small
\setlength{\parindent}{0pt}
An infinite sequence $a_1$, $a_2$, \dots\ consists of positive integers \newline
has each of which has at least three proper divisors. \newline
Suppose that for each $n\geq 1$, \newline
$a_{n+1}$ is the sum of the three largest proper divisors of $a_n$. \newline
Determine all possible values of $a_1$.
\endgroup
\end{promptbox}
\begin{evalbox}{Human Evaluation: QED-Nano (0/7)}
\begingroup
\small
\setlength{\parindent}{0pt}
While the model correctly argues that any term in the sequence cannot be uneven, its proof is strange and incorrect. \newline
In particular, it uses two identical divisors in its three divisors rather than three distinct ones. \newline
Its proof of Lemma 3 is also incorrect, by assuming that $3 \\not\\mid m$. \newline
It considers the case $m$ prime, but by the problem statement this never occurs. \newline
The LLM judge is too lenient for these mistakes. \newline
The proof clearly shows a lack of understanding and should be given 0 points.
\endgroup
\end{evalbox}
\begin{evalbox}{Human Evaluation: QED-Nano (Agent) (5/7)}
\begingroup
\small
\setlength{\parindent}{0pt}
The model makes a very strange mistake by stating that $\frac{26}{20}m < m$. \newline
However, the error seems a simple typo: $\frac{1}{2} + \frac{1}{4} + \frac{1}{5} = \frac{19}{20}$, not $\frac{26}{20}$. \newline
The model also fails to prove that if $m$ is not divisible by $3$, $f(m)$ is not divisible by $3$. \newline
We deduct two points for this.
\endgroup
\end{evalbox}

\subsection{IMO-2025-Q5}
\begin{promptbox}{Problem}
\begingroup
\small
\setlength{\parindent}{0pt}
Alice and Bazza are playing the inekoalaty game, \newline
a two-player game whose rules depend on a positive real number $\lambda$ which is known to both players. \newline
On the $n$th turn of the game (starting with $n=1$) the following happens:

  (ii) If $n$ is odd, Alice chooses a nonnegative real number $x_n$ such that \newline
  \[ x_1 + x_2 + \cdots + x_n \le \lambda n.  \] \newline
  (ii) If $n$ is even, Bazza chooses a nonnegative real number $x_n$ such that \newline
  \[ x_1^2 + x_2^2 + \cdots + x_n^2 \le n.  \]

If a player cannot choose a suitable $x_n$, the game ends and the other player wins. \newline
If the game goes on forever, neither player wins. \newline
All chosen numbers are known to both players.

Determine all values of $\lambda$ for which Alice has a winning strategy \newline
and all those for which Bazza has a winning strategy.
\endgroup
\end{promptbox}
\begin{evalbox}{Human Evaluation: QED-Nano (0/7)}
\begingroup
\small
\setlength{\parindent}{0pt}
The proof is completely incorrect. The LLM judge is once again too lenient in its grading. \newline
No partial credits should be given for an optimal strategy on a subset of the values of $\lambda$.
\endgroup
\end{evalbox}
\begin{evalbox}{Human Evaluation: QED-Nano (Agent) (7/7)}
\begingroup
\small
\setlength{\parindent}{0pt}
The argument is correct and well-structured. A bit too verbose, but not to the extent where it gets annoying.
\endgroup
\end{evalbox}

\subsection{IMO-2025-Q6}
\begin{promptbox}{Problem}
\begingroup
\small
\setlength{\parindent}{0pt}
Consider a $2025\times2025$ grid of unit squares. \newline
Matilda wishes to place on the grid some rectangular tiles, \newline
possibly of different sizes, \newline
such that each side of every tile lies on a grid line \newline
and every unit square is covered by at most one tile.

Determine the minimum number of tiles Matilda needs to place \newline
so that each row and each column of the grid has exactly one unit square \newline
that is not covered by any tile.
\endgroup
\end{promptbox}
\begin{evalbox}{Human Evaluation: QED-Nano (0/7)}
\begingroup
\small
\setlength{\parindent}{0pt}
While the grading scheme created by ProofBench says that a reformulation of the problem in terms of permutations should earn one point, \newline
this is far too trivial to actually warrant the point. This is the only thing the model does correctly (as expected for this difficult problem) \newline
The grading scheme in MathArena is stricter, and does not award a point for this trivial step.
\endgroup
\end{evalbox}
\begin{evalbox}{Human Evaluation: QED-Nano (Agent) (0/7)}
\begingroup
\small
\setlength{\parindent}{0pt}
While the grading scheme created by ProofBench says that a reformulation of the problem in terms of permutations should earn one point, \newline
this is far too trivial to actually warrant the point. This is the only thing the model does correctly (as expected for this difficult problem) \newline
The grading scheme in MathArena is stricter, and does not award a point for this trivial step.
\endgroup
\end{evalbox}




\end{document}